\documentclass[sigconf,screen]{acmart}
\usepackage{graphicx}
\usepackage{caption}
\usepackage{subcaption}
\usepackage{microtype}
\usepackage{multirow}
\graphicspath{ {./figures/} }
\usepackage{tabularx}
\usepackage{array}
\usepackage{balance}

\usepackage[normalem]{ulem}

\setcopyright{acmlicensed}
\acmPrice{15.00}
\acmDOI{10.1145/3503222.3507777}
\acmYear{2022}
\copyrightyear{2022}
\acmSubmissionID{asplos22main-p1386-p}
\acmISBN{978-1-4503-9205-1/22/02}
\acmConference[ASPLOS '22]{Proceedings of the 27th ACM International Conference on Architectural Support for Programming Languages and Operating Systems}{February 28 -- March 4, 2022}{Lausanne, Switzerland}
\acmBooktitle{Proceedings of the 27th ACM International Conference on Architectural Support for Programming Languages and Operating Systems (ASPLOS '22), February 28 -- March 4, 2022, Lausanne, Switzerland}

\begin{document}

\title[RecShard]{RecShard: Statistical Feature-Based Memory Optimization \\ for Industry-Scale Neural Recommendation}

\author{Geet Sethi}
\affiliation{%
  \institution{Stanford University and Meta}
  \city{Stanford}
  \state{California}
  \country{USA}}
\email{geet@cs.stanford.edu}

\author{Bilge Acun}
\affiliation{%
  \institution{Meta}
  \city{Menlo Park}
  \state{California}
  \country{USA}}
\email{acun@fb.com}

\author{Niket Agarwal}
\affiliation{%
  \institution{Meta}
  \city{Menlo Park}
  \state{California}
  \country{USA}}
\email{niketa@fb.com}

\author{Christos Kozyrakis}
\affiliation{%
  \institution{Stanford University}
  \city{Stanford}
  \state{California}
  \country{USA}}
\email{kozyraki@stanford.edu}

\author{Caroline Trippel}
\affiliation{%
  \institution{Stanford University}
  \city{Stanford}
  \state{California}
  \country{USA}}
\email{trippel@stanford.edu}

\author{Carole-Jean Wu}
\affiliation{%
  \institution{Meta}
  \city{Cambridge}
  \state{Massachusetts}
  \country{USA}}
\email{carolejeanwu@fb.com}

\begin{abstract}
We propose RecShard, a fine-grained embedding table (EMB) partitioning and placement technique for deep learning recommendation models (DLRMs). RecShard is designed based on two key observations. First, not all EMBs are equal, nor all rows within an EMB are equal in terms of access patterns. EMBs exhibit distinct memory characteristics, providing performance optimization opportunities for intelligent EMB partitioning and placement  across a tiered memory hierarchy. Second, in modern DLRMs, EMBs function as hash tables. As a result, EMBs display interesting phenomena, such as the birthday paradox, leaving EMBs severely under-utilized. RecShard determines an optimal EMB sharding strategy for a set of EMBs based on training data distributions and model characteristics, along with the bandwidth characteristics of the underlying tiered memory hierarchy. In doing so, RecShard achieves over 6 times higher EMB training throughput on average for capacity constrained DLRMs. The throughput increase comes from improved EMB load balance by over 12 times and from the reduced access to the slower memory by over 87 times.
\end{abstract}




\maketitle

\section{Introduction}
\label{sec:intro}

Deep learning (DL) is pervasive, supporting a wide variety of application domains~\cite{imagenet,devlin-etal-2019-bert,hazelwood:2018:mlatfb,resnet,AlphaFold,naumov2020deep,transformers,Weyn_2020}.
A significant fraction of deep learning compute cycles in industry-scale data centers can be attributed to deep learning recommendation models (DLRMs)~\cite{Cheng:dlrs2016, covington:2016:youtuberec, netflix,gupta:2020:archimp, tpu, Raimond:netflix, what_to_watch, zhou2019deep,wu:arxiv:2021}.
For example, at Facebook, DLRMs
account for more than 50\% of training demand~\cite{naumov2020deep} and more than 80\% of inference demand~\cite{gupta:2020:archimp}.
Moreover, Google's search engine relies on its recommender system, such as RankBrain, for search query processing~\cite{googe:2016:rankbrain}.

\noindent \textbf{DLRMs} \quad
DLRMs exhibit distinct systems implications compared to more traditional neural network architectures~\cite{deeprecsys,rec-nmp,kumar2021exploring,wang2020exploiting,hsia:iiswc:2020}. This is due to their use of \textit{embedding layers} which demand orders-of-magnitude higher memory capacity and exhibit significantly lower compute-intensity~\cite{gupta:2020:archimp,naumov2019deep,lui:2021:capacity}.
Embedding layers, comprised of \textit{embedding tables} (EMBs), support the transformation of categorical (i.e., sparse) features into dense representations.
Categorical features are typically represented as one-hot or multi-hot binary vectors, where entries represent feature categories.
Activated categories (binary value of 1) in a feature vector then induce a set of look-ups to the feature's corresponding EMB to extract dense latent vectors. 

\begin{figure}[t]
\subfloat[DLRM memory requirements have grown by 16x, while memory capacity on GPU accelerators has improved by less than 6x.]{%
\centering
  \includegraphics[clip,width=.48\linewidth, ,keepaspectratio]{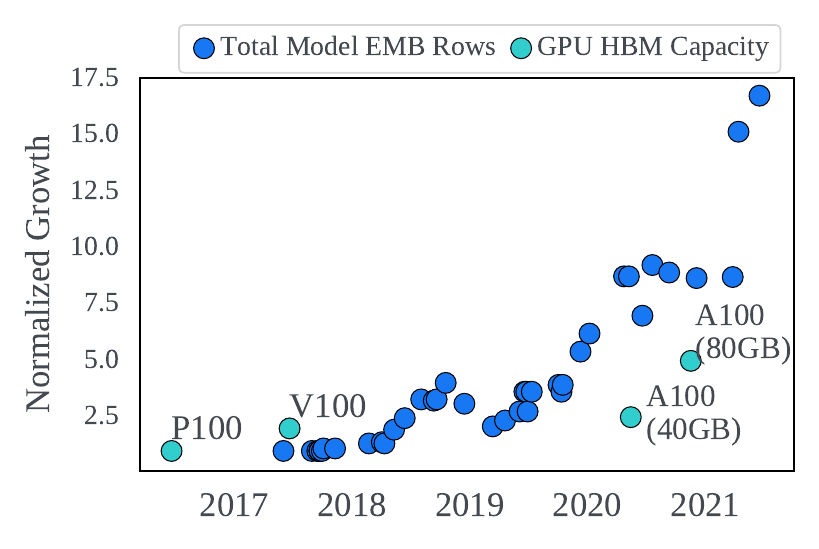}%
  \label{fig:opening-a}
}
\hspace{1mm}
\subfloat[DLRM bandwidth demands have grown by ~30x, far outpacing the bandwidth growth of accelerator memories and interconnects.]{%
\centering
  \includegraphics[clip,width=.48\linewidth, ,keepaspectratio]{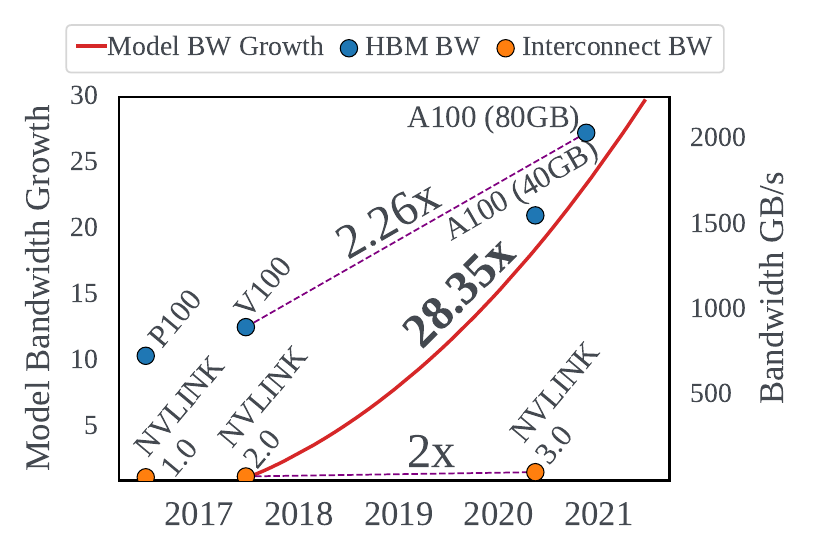}%
  \label{fig:opening-b}
}

\caption{DLRM system requirement growth trend.
}
\label{fig:opening}
\end{figure}

\noindent \textbf{System Requirement Characteristics for DLRMs} \quad
The large feature space for industry-scale DLRMs demands significant compute throughput (PF/s), memory capacity (10s of TBs), and memory bandwidth (100s of TB/s)~\cite{mudigere2021highperformance}.
Figure~\ref{fig:opening} illustrates that the \textit{memory capacity and bandwidth demands for DLRMs have been growing super-linearly, exceeding the memory capacities available on training hardware.}
Figure~\ref{fig:opening-a} shows that between 2017-2021, the memory capacity requirements of DLRMs have grown by 16 times.
EMB memory footprints are on the order of terabytes (TB)~\cite{lui:2021:capacity,zhao:2020:distributedgpu} and account for over 99\% of the total model capacity~\cite{gupta:2020:archimp}.
The growth in the number and sizes of EMBs stems from the increase in the number of features and feature categories represented, in order to improve the overall DLRM prediction quality.
Figure~\ref{fig:opening-b} shows that, within the same four-year period, per-sample DLRM memory bandwidth demand, determined by the amount of EMB rows accessed in a single training data sample, has increased by almost 30 times, outpacing the growth and availability of memory bandwidth on state-of-the-art training hardware.

\noindent \textbf{Hierarchical Memory in Training Systems} \quad ~The widening gap between the DLRM memory needs and the memory specifications of modern training system hardware motivates new memory optimization techniques to effectively scale training throughput.
While the exact training system architectures differ, hierarchical memory systems, e.g. tiered hierarchies composed of GPU HBM, CPU DRAM, and SSD~\cite{zhao:2020:distributedgpu}, are becoming increasingly common for DLRM training.
Since not all EMBs can fit entirely in GPU HBMs, this scenario gives rise to optimization strategies to address the first challenge -- deciding \textit{where EMBs should be placed in the hierarchical memory system} to maximize training throughput.
Strategically placing EMBs brings up the second challenge -- ensuring \textit{efficient utilization of all available memory capacity and bandwidth}.

\noindent \textbf{Characterizing EMB Access Patterns for DLRMs} \quad
In this paper, we make two key observations regarding the memory access behaviors of EMBs that motivate more performant and efficient EMB partitioning and placement schemes.

\begin{figure}[t]
  \centering
    \includegraphics[width=0.85\linewidth,keepaspectratio]{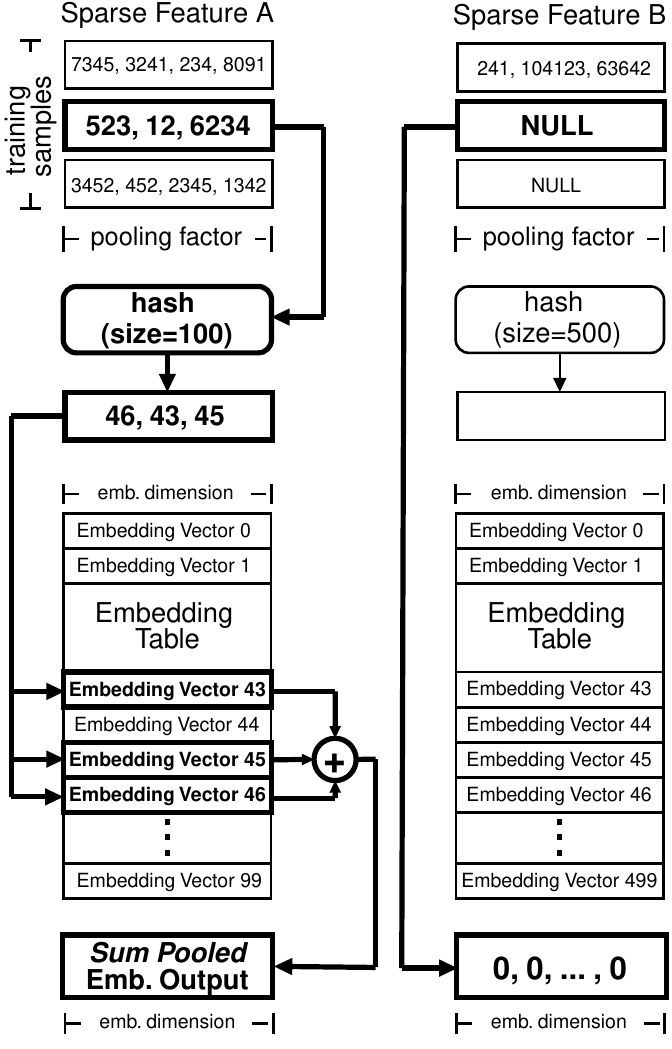}
    \caption{Example illustrating the pooling factor and coverage statistics, along with the embedding lookup and (sum) pooling operation. In this example there are two sparse features, \textit{A} and \textit{B}, with two corresponding embedding tables, and a training dataset composed of three training data samples. The average pooling factors of sparse features \textit{A} and \textit{B} over the dataset are 3.66 and 3, respectively, while the coverages are 1.0 and .33, respectively. The example shows the embedding lookup and pooling operation for the second training data sample (highlighted in bold). For sparse feature \textit{A}, the raw input IDs are \textit{hashed} with an output size of 100 (which corresponds to the number of rows in \textit{A's} EMB), generating the corresponding embedding lookup indices. These embedding rows, each containing \textit{embedding dimension} number of values, are then read and combined, i.e. \textit{pooled}, via element-wise summation to produce the output vector of the lookup operation. For sparse feature \textit{B}, the second training data sample is \textit{NULL}, signifying that \textit{B} contains no feature data for that particular data sample. This results in the stages which sparse feature \textit{A} went through being bypassed and a 0-vector being produced as the output. 
    }
  \label{fig:emb_example}
\end{figure}

First, not all EMBs are equal, nor are all rows within an EMB equal in terms of access behaviors.
For example, the \textit{frequency distribution} of a sparse feature's categorical values often follows a power law distribution. Therefore, a relatively small fraction of EMB rows will source the majority of all EMB accesses. 
Furthermore, as illustrated in Figure~\ref{fig:emb_example}, sparse features, and thus EMBs, exhibit varying bandwidth demands due to varying \textit{pooling factors} -- the number of activated categories on average in a particular sparse feature sample -- and \textit{coverage} -- the fraction of training samples in which a particular feature appears.
Second, in modern DLRMs, EMBs function as \textit{hash tables}. As a result, EMBs display interesting phenomena, such as the birthday paradox, which leaves a significant portion of EMBs unused due to hash collisions. Unused EMB space is further increased with increasing \textit{hash sizes}.

Building on the in-depth sparse feature characterization of production scale DLRMs (Section~\ref{sec:characterization}),
we propose \textit{RecShard} -- 
a new approach to improve DLRM training throughput using a \textit{data-driven and system-aware EMB partitioning and placement strategy}.
RecShard's EMB \textit{sharding} strategy is informed by per-feature training data distributions---categorical value frequency distributions (Figure~\ref{fig:freq-cdf}), pooling factor statistics (Figure~\ref{fig:pooling}) as well as coverage distributions of all sparse features (Figure~\ref{fig:coverage}). 
RecShard also considers EMB design settings---hash functions and table sizes (Figure~\ref{fig:sparsity}) as well as characteristics of the underlying tiered memory.
RecShard considers the training system design parameters simultaneously through the use of a mixed integer linear program (MILP) to produce an optimal EMB sharding strategy. 
Overall, the key contributions of this paper are as follows:

\begin{itemize}
    \item \textbf{Fine-grained, data-driven EMB sharding:} \quad We demonstrate that
    EMB access patterns during DLRM training vary within and across EMBs. As a result, DLRM training throughput stands to improve with fine-grained EMB sharding. Further,  EMB access patterns can be estimated by deriving statistics from less than 1\% of training data (categorical value frequency distribution, pooling factor, and coverage) and the target DLRM architecture (hash function and hash size).
    Thus, intelligent EMB sharding schemes can be instituted prior to training time. %
    \item \textbf{RecShard:} \quad We propose RecShard -- a new approach for fine-grained sharding of EMBs with respect to a multi-level memory hierarchy consisting of GPU HBM and CPU DRAM. RecShard optimizes EMB partitioning and placement globally based on the estimated sparse feature characteristics and DLRM architecture. %
    \item \textbf{Real system evaluation:} \quad To demonstrate its efficacy, we implement and evaluate RecShard in the context of a production scale DLRM. We demonstrate that RecShard can on average improve the performance and load balance of DLRM EMB training by over 5x and over 11x, respectively, compared to the state-of-the-art industry sharding strategies~\cite{acun:2021:understandingtraining, lui:2021:capacity,mudigere2021highperformance}.
\end{itemize}

\begin{figure}[t]
  \centering
    \includegraphics[width=\linewidth,keepaspectratio]{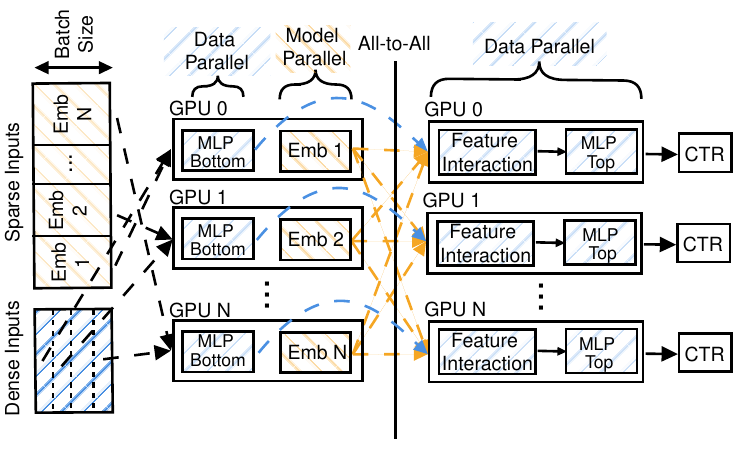}
    \caption{Generalized hybrid-parallel DLRM architecture. Data parallel modules (MLPs) are shaded blue while model parallel EMBs are shaded orange.}
  \label{fig:dlrm_arch}
\end{figure}

\section{Background}
\label{sec:back}

Figure~\ref{fig:dlrm_arch} gives an overview of the canonical Deep Learning Recommendation Model (DLRM) architecture~\cite{naumov2019deep}. In this section, we provide background on DRLMs and the training systems.%

DLRMs process user-content pairs to predict the probability that a user will interact with a particular piece of content, commonly referred to as the click-through-rate (CTR). To produce such a prediction, DLRMs consume two types of features: \emph{dense} and \emph{sparse}. Dense features represent continuous data, such as a user's age or the time of day, while sparse features represent categorical data, such as domain names or recent web pages viewed by a user. To encode this categorical data, sparse features are represented as one-hot or multi-hot binary vectors which are only activated for a small subset of relevant categories (hence the term \emph{sparse}).
Sparse features used in DLRMs can have cardinalities in the billions~\cite{kang2021learning, zhao:2020:distributedgpu}.  

At a high level, the primary components of DLRMs are Multi-Layer Perceptrons (MLPs) and Embedding Tables (EMBs). 
EMBs are commonly-used to transform sparse features from the high-dimensional, sparse input space to low-dimensional, dense embedding vectors.
EMBs perform this operation by functioning as large lookup tables, where, in theory, each rows acts as a latent vector encoding of a particular sparse feature value (i.e., category). The activated, or hot, indices of the sparse inputs then act as indices into the EMBs, gathering one or more embedding vectors. 

In practice, however, the binary-encoded sparse feature inputs are
hashed prior to EMB look-up. Hashing serves two purposes. First, hashing allows the bounding of a sparse feature's EMB to a pre-determined, fixed size. Second, hashing permits the handling of unseen inputs at runtime~\cite{acun:2021:understandingtraining, kang2021learning}. Once gathered, the embedding vectors are aggregated on a per-EMB basis using a \emph{feature pooling} operation, such as summation or concatenation. The pooled vectors, along with the outputs of the bottom MLP layers (which process dense inputs), are then combined using a \emph{feature interaction} layer, before proceeding through the top MLP layers and producing a prediction for the estimated engagement for the user-content pair.     

\begin{figure}[t]
  \centering
    \includegraphics[width=.90\linewidth,keepaspectratio]{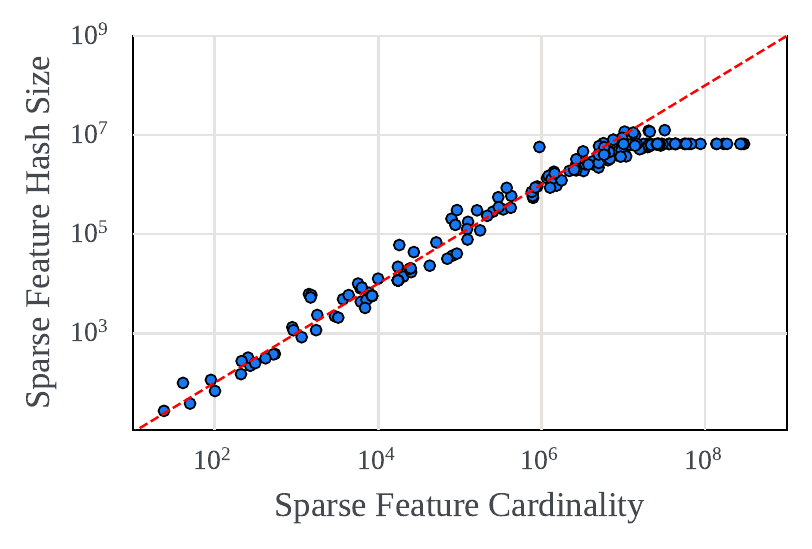}
    \caption{Sparse feature cardinality (categorical space; x-axis) versus chosen feature hash size (EMB size; y-axis) for 200 sparse features used in a large production-scale model. Hash size equal to cardinality is shown by the red-dotted line.}
  \label{fig:model_card_hash}
\end{figure}

\noindent \textbf{Training Systems for DLRMs} \quad
\label{sec:training-systems}
DLRMs present significant infrastructure challenges. While the MLP layers are compute-intensive and exhibit (relatively) small memory footprints, single EMBs of production-scale DLRMs can be on the order of 100s of gigabytes each, with the total memory capacity on the multi-TB scale~\cite{kang2021learning, mudigere2021highperformance, zhao:2020:distributedgpu}. 
Furthermore, EMBs exhibit irregular memory access patterns~\cite{wilkening:2021:recssd}, and the concurrent vector accesses per-EMB and across EMBs require substantial memory bandwidth~\cite{acun:2021:understandingtraining, rec-nmp}. This has led to a hybrid data- and model-parallel training approach (Figure~\ref{fig:dlrm_arch}). MLP layers (both top and bottom) are replicated across all trainers (GPUs in figure) in a data-parallel manner, while EMBs are sharded across trainers to exploit model-parallelism~\cite{xdl, unified_arch_osdi, mudigere2021highperformance, yin:2021:ttrec}.

The ever-increasing memory capacity and bandwidth demands of DLRM training has also led to the emergence of training systems with tiered hierarchical memories (such as hierarchies with HBM, DRAM, and SSD tiers). The large collection of EMBs are partitioned and/or cached across the various tiers~\cite{mudigere2021highperformance,zhao:2020:distributedgpu}. One class of partitioning approaches leverages \emph{unified virtual memory} (UVM)~\cite{nvidia:2013:uvm}. This places both host DRAM and accelerator HBM in a shared virtual address space, allowing transparent access of host DRAM on a GPU accelerator without explicit host-device transfers~\cite{pump_up_sigmod, emogi}. UVM can greatly expand the usable memory capacity of a GPU node with ease. For example, a server with 8x 32GB HBM GPUs can have 2TB of DRAM~\cite{acun:2021:understandingtraining}. 

However, for memory-bound workloads, such as DLRMs, using UVM naïvely can come with significant performance cost. While the latest GPUs contain HBMs with bandwidth capacity approaching 2TB/s, the interconnects used can have bandwidth capacity an order of magnitude less. Single direction throughput of PCIe 4.0x16, for example, is just 32 GB/s. This places particular importance on the DLRM EMB sharding scheme---\textit{hundreds of EMBs with heterogeneous memory characteristics have to be placed across potentially hundreds of trainers.}

To address the performance needs of production-scale DLRM training in the presence of rapidly-growing memory capacity and bandwidth demands, this paper focuses on the \emph{partitioning and placement problem}---determining the optimal placement of EMBs on a tiered memory system with fixed memory capacity and bandwidth constraints.

\section{Characterization of DLRM Sparse Features}
\label{sec:characterization}

The goal of a \textit{DLRM sharder} is to partition a model's EMBs across a training system's hardware topology, in order to fully exploit model parallelism and thereby maximize training throughput. This requires an EMB placement across an increasingly tiered memory hierarchy that balances training load across all trainers (GPUs).
To achieve such load balancing, an effective EMB sharder must be able to accurately estimate the memory demands of each EMB.
RecShard addresses this problem through a data-driven approach.

This section presents our in-depth memory characterization of sparse features used in industry-scale DLRMs. The characterization study captures the statistical nature of recommendation training data,  
and sheds light on five key characteristics of DLRM sparse features which RecShard exploits to improve the EMB training throughput performance. Notably, we find that a sparse feature’s \textit{value distribution} enables us to determine the portion of an EMB that will exhibit high temporal locality during training, the feature’s \textit{average pooling factor} provides a proxy for its memory bandwidth cost, and the feature’s \textit{coverage} allows us to rank the placement priorities across EMBs.
Furthermore, these statistics are \textit{distinct and vary over time} for each sparse feature.

\subsection{Skewed Categorical Distribution Presents Unique EMB Locality Characteristics}
\label{sec:observation-1}

\textit{{\large A small subset of categories can constitute the majority of accesses to an EMB.}}

\begin{figure}[t]
  \centering
    \includegraphics[width=.90\linewidth]{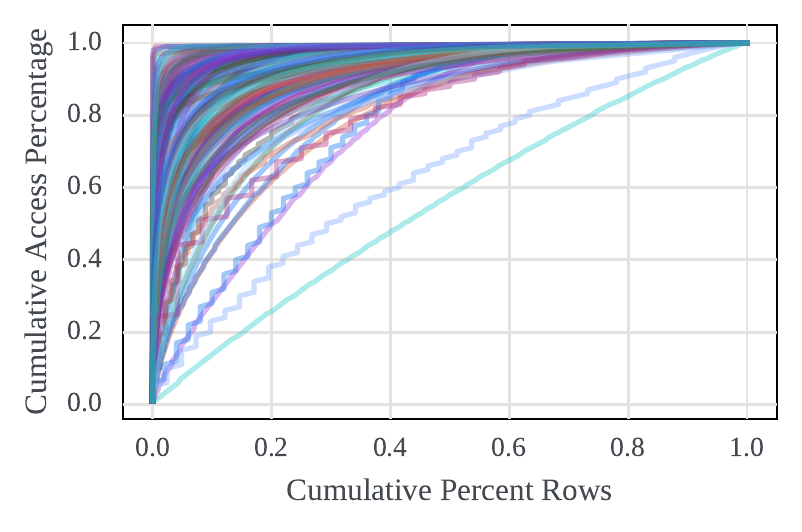}
    \caption{Hashed Value Frequency CDFs of 200 sparse features used in a production DLRM. The CDFs are generated from over two billion randomly-selected training samples over ten days of data, \emph{post-hash}.}
  \label{fig:freq-cdf}
\end{figure}

Sparse features represent categorical data, with each sparse feature's data sample containing a variable length list of categorical values from its sparse feature space. As the size of this categorical feature space can be arbitrarily large, it is natural to ask if a subset of values appear more often than others, and in fact they do~\cite{ginart2019mixed, Joglekar:2020, Liu:2020, wu:2020:mlperfrec}.
For example, the country a user is located in is a common feature for recommendation use cases. If we were to measure the distribution of this feature, we would see the feature follows a skewed power-law distribution, as the world population by country itself follows a power-law distribution with a long tail. Production-scale DLRMs often consist of hundreds of features that exhibit similar categorical frequency distributions~\cite{acun:2021:understandingtraining, mudigere2021highperformance}.

Figure~\ref{fig:freq-cdf} illustrates the cumulative distribution function (CDF) of 200 representative categorical features of a production DLRM.
While a handful of features exhibit more uniform value distributions, the vast majority display a power-law distribution over the categorical values. In other words, for the majority of features, a small subset of categories appear much more frequently than the rest. This implies that a small set of EMB rows comprise the majority of EMB accesses. It is important to also highlight that \textit{the strength of the distribution varies from one feature to another, requiring consideration of the distribution on a per-feature basis}. 

Overall, the locality characteristics unique to each feature give rise to an optimization opportunity -- EMB entries \textit{within a table} can be placed across a tiered memory hierarchy based on expected access patterns. We refer to this optimization as fine-grained EMB partitioning.

\subsection{Pooling Factors Determine Memory Bandwidth Demand}
\label{sec:observation-2}
\textit{{\large Within a training data sample, each EMB exhibits its own bandwidth demand due to varying pooling factor distributions.}}

\begin{figure}[t]
\subfloat[\textit{Average pooling factor}: the number of `hot' indices in an average sparse feature's input sample.]{%
\centering
  \includegraphics[clip,width=.48\linewidth, ,keepaspectratio]{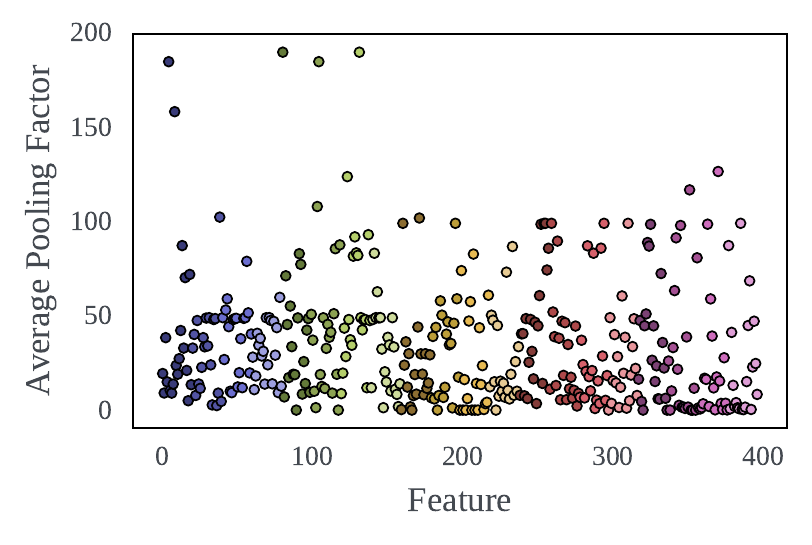}%
  \label{fig:pooling}
}
\hspace{1mm}
\subfloat[\textit{Coverage percentage}: the probability a sparse feature is present in a random training data sample.]{%
\centering
  \includegraphics[clip,width=.48\linewidth, ,keepaspectratio]{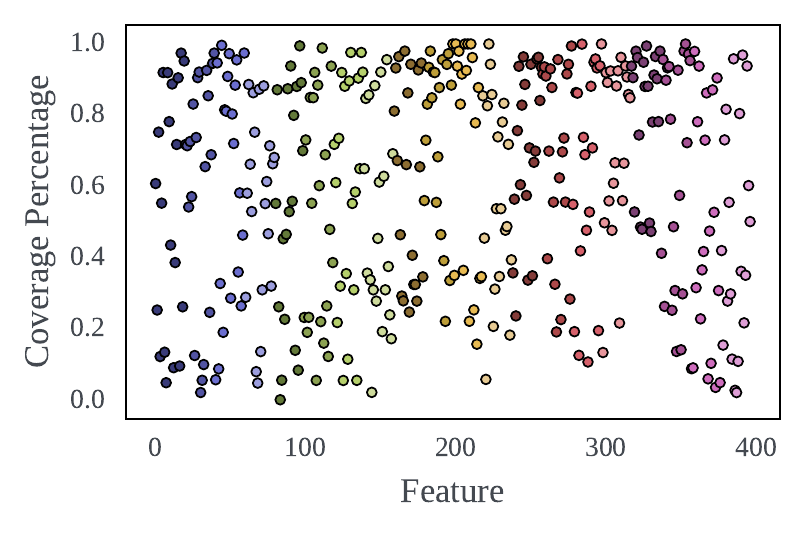}%
  \label{fig:coverage}
}

\caption{Average pooling factor
and coverage
vary widely from feature to feature. Collectively, they serve as a proxy for the per-sample bandwidth demand of a feature.
}
\vspace{-.5cm}
\label{fig:random}
\end{figure}

Activated indices in a sparse feature's input effectively correspond to the rows in the feature's EMB that should be accessed to acquire latent vector representations of the categories. This results in a scatter-gather memory access pattern, where one embedding vector is accessed for each activated index. 
The $n$ EMB rows accessed by a sparse feature's input is its sample \emph{pooling factor}, whereas the interaction of the corresponding $n$ latent embedding vectors via \emph{pooling} determines the feature sample's representation. The distribution of the pooling factors -- $n$ -- of a sparse feature across the training data models the feature's memory bandwidth consumption.

Furthermore, 
the pooling factor distribution 
can vary from feature to feature, resulting in memory bandwidth needs that are feature-specific (i.e., EMB-specific).
This is due to variability in the information each feature represents. While the feature representing the location of a user may always be of length one, a feature representing the pages recently viewed by a user will likely have length greater than one.  
Figure~\ref{fig:pooling} depicts the average pooling factor distribution for hundreds of sparse features which varies widely. Some sparse features exhibit high pooling factors of approximately two hundred on average, while the average pooling factors of others are on the order of a few tens; the result is an order of magnitude difference in the memory bandwidth demand. 

As with sparse feature value distributions, the pooling distributions for sparse features are also skewed with a long tail; however unlike the value distributions, they cannot be broadly classified as being power-laws with varying degrees of strengths. We experimented with an assortment of summary statistics, such as the median and mean, to determine which provides the best estimate for the `average' case across all features; resulting in the choice of mean as the estimate for the \emph{average pooling factor} of a sparse feature. This choice was made as we observed that the mean generally tends to over-estimate an EMB's bandwidth demand, which we find preferable to under-estimating and potentially resulting in a sub-optimal EMB placement.

In summary, pooling factor diversity across features motivates optimizations that consider per-feature \emph{average pooling factors} to approximate the unique memory bandwidth consumption characteristics for EMBs.

\subsection{Varying Degrees of Coverage for Sparse Features Determines EMB Placement Priority}
\label{sec:observation-3}

\textit{{\large Sparse features exhibit varying degrees of coverage, with some EMBs being used much more often than others.}}

Not all sparse features of a DLRM are referenced in each training data sample. 
There are a variety of reasons for this, such as a particular feature being phased in or out, or a user simply not having the content interaction or metadata necessary for the feature to be instantiated. Regardless of the reason,
there is variability in the presence of sparse features across training inputs, which provides us with additional empirical information for system performance optimizations.

Figure~\ref{fig:coverage} depicts the feature access probabilities (y-axis) across hundreds of sparse features sampled from a number of industry-scale DLRMs (x-axis).  
The probability that a sparse feature is present in a training sample is referred to as its \emph{coverage}. 
Similar to the pooling factor distribution (Section~\ref{sec:observation-2}), the coverage of individual sparse features varies widely from feature to feature -- ranging from less than 1\% on the low-end to 100\% on the high-end. This observation demonstrates the importance of considering per-feature coverage characteristics in EMB placement decisions. %
Thus, a feature’s \textit{coverage} gives rise to system optimizations based on the prioritization of EMBs according to their frequency of use. 

\subsection{Embedding Hashing Leads to Sub-optimal System Memory Utilization}
\label{sec:observation-4}

\textit{{\large While a simple technique, embedding hashing is inefficient from the perspective of system memory utilization.}}

\begin{figure}[t]
  \centering
    \includegraphics[width=.90\linewidth,keepaspectratio]{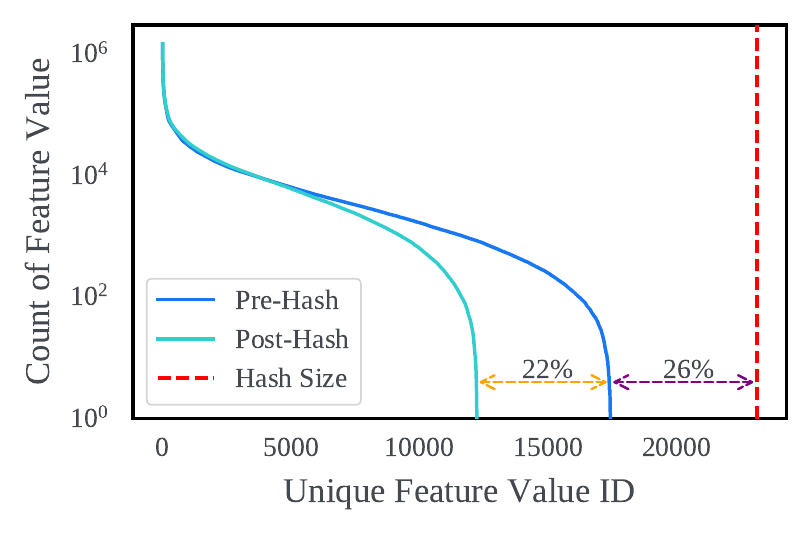}
    \caption{The impact of hashing on the feature value frequency distribution. Even using a hash size greater than the number of unique values, hashing causes the compression of the raw value distribution, leaving considerable EMB under-utilization.}
  \label{fig:sparsity}
\end{figure}

The cardinality of a given sparse feature can be on the order of billions. Thus, constructing an EMB representing the entirety of such a sparse feature would be prohibitively expensive in terms of the memory capacity requirement. Furthermore, it would not generalize to unseen feature values when new categories emerge. Thus, it is unrealistic to construct a one-to-one mapping between every sparse feature value and EMB rows. Instead, the EMBs of industry-scale DLRMs typically employ hashing~\cite{acun:2021:understandingtraining, kang2021learning, hashing_trick}, using a random hash function to map arbitrary feature values to output values constrained by a specified hash size. The hash size therefore dictates the size of the EMB.

A consequence of using random hashing to map a feature's inputs to corresponding EMB entries is \emph{hash collisions}---where the hash function maps two unique input values to the same output value. The existence of hash collisions can be demonstrated via the pigeonhole principle, as mapping $H+1$ unique values with a hash size of $H$ requires at least two input value overlap. What is less obvious however, is whether or not, and to what degree, collisions occurs when the hash size is equal to or even slightly greater than the number of unique input values seen. 
Commonly known as the birthday paradox, when hashing $N$ unique input values with a hash size of $H=N$, one will observe that approximately $\frac{1}{e}$ input values will collide. And, as 
$N=H$, this results in $\frac{1}{e}$ hash entries being unused. 

Figure~\ref{fig:sparsity} depicts the birthday paradox phenomenon by illustrating the pre- and post-hash distributions for a specific feature of a production DLRM. The pre-hash distribution (dark blue line) depicts the input feature value space, whereas the post-hash distribution (light blue line) depicts the distribution of accesses to the corresponding EMB. The red-dotted vertical line denotes the specified hash size and therefore the number of unique embedding vectors that can be captured by this EMB.
Although the hash size is greater than the number of unique pre-hash values observed (the red dotted line is to the \emph{right} of the dark blue line), 
the post-hash embedding space compresses the pre-hash categorical feature space (the light blue line terminates to the \emph{left} of the dark blue line). Furthermore, Figure~\ref{fig:sparsity} highlights the under-utilization of EMBs due to training data sparsity by 26\% and hash collisions by another 22\%.

\begin{figure}[t]
  \centering
    \includegraphics[width=.90\linewidth,keepaspectratio]{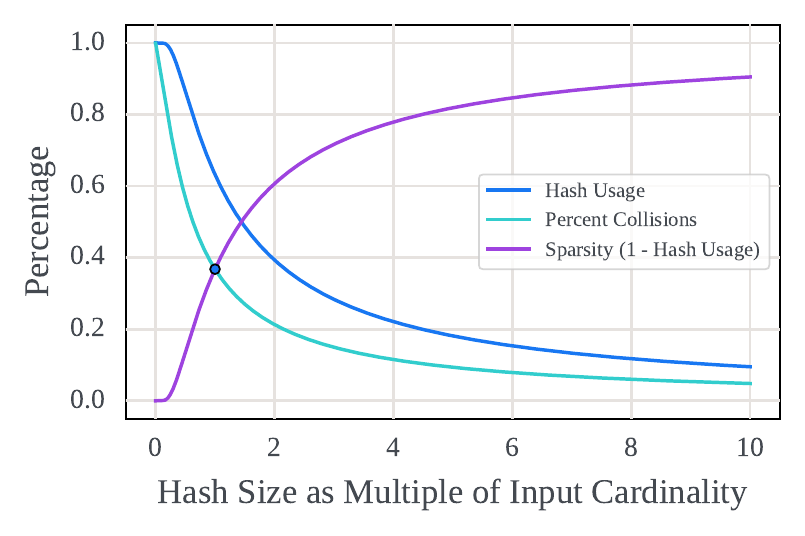}
    \caption{Increasing the hash size to accommodate the tail leaves an increasing percentage of the hash space unused, which RecShard can reclaim. The blue dot denotes the point at which hash size is equal to input cardinality.}
    
  \label{fig:hashing-tail}
\end{figure}

Increasing the hash size to accommodate the tail of the power-law distribution -- a technique which can improve model performance~\cite{zhao:2020:distributedgpu} -- leaves an increasing percentage of the hash space under-utilized, which RecShard can reclaim. 
Figure~\ref{fig:hashing-tail} illustrates %
that, as the hash size is increased to accommodate the tail of the input sparse feature distribution (Section~\ref{sec:observation-1}), an increasing percentage of the hash space is unused by training samples (sparsity increases). 

Given the observations above, hashing
gives additional insight into designing an intelligent partitioning strategy for EMBs. Due to the birthday paradox and the desire to choose a hash size which can retain as much of the tail as possible, a non-trivial percentage of embedding rows will not be accessed at all during training. This enables us to move the under-utilized portions of EMBs to a slower memory tier (or potentially avoid allocation altogether) without visible impact on the training time performance.

\begin{figure}[t]
  \centering
    \includegraphics[width=.90\linewidth,keepaspectratio]{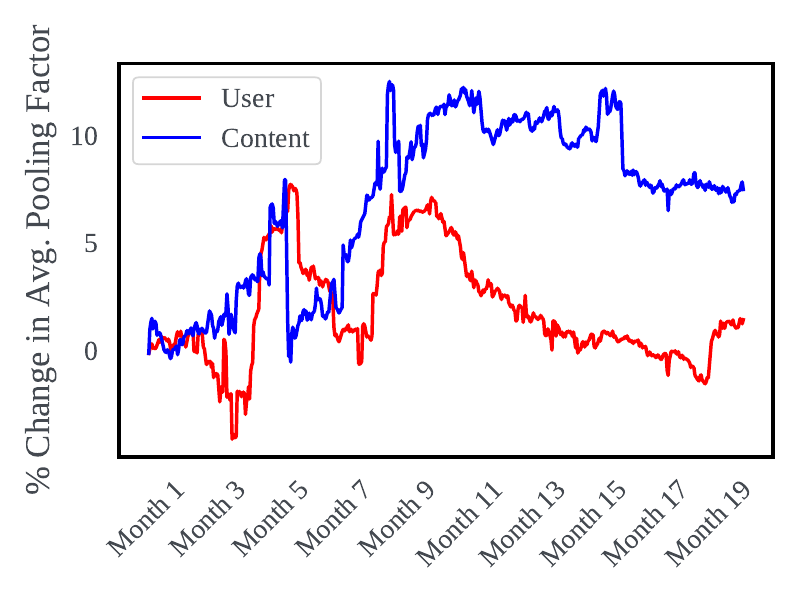}
    \caption{Sparse features are grouped into two general categories, users and content. Both feature types exhibit dynamic memory demand over time. We show memory demand for a large production model (\char`\~400 features) over a 20-month period. Data represent averages over all relevant features.}
  \label{fig:features-over-time}
\end{figure}

\subsection{Sparse Feature Memory Patterns Evolve over Time}
\label{sec:observation-5}

\textit{{\large Sparse features exhibit distinct, dynamic memory demands over time.}}

Sections~\ref{sec:observation-1}-\ref{sec:observation-4} provide insights into \textit{how} memory characteristics specific to DLRM sparse features and EMB design can be used to optimize the EMB performance of DLRMs through an intelligent data-driven sharding strategy. It is, however, also important to know \emph{how often} EMB sharding should be performed.
Once deployed, industry-scale production models may be continuously retrained on new data for potentially many weeks~\cite{hazelwood:2018:mlatfb} at a time.

Figure~\ref{fig:features-over-time} illustrates how average feature lengths evolve over a 20-month time period for two distinct types of sparse features: \textit{content} features and  \textit{user} features. Based on the time-varying nature of sparse feature statistics, ideally the benefit of re-sharding would be evaluated regularly throughout training as new data arrives, due to the potentially large impact that data distribution shifts can have on training throughput. Although this benefit can be approximated quickly by RecShard (Section~\ref{sec:recshard}), it must be dynamically weighed against the cost of carrying out the re-sharding on the given training stack and topology.

\section{RecShard}
\label{sec:recshard}

\begin{figure*}[t]
  \centering
    \includegraphics[scale=.9,keepaspectratio]{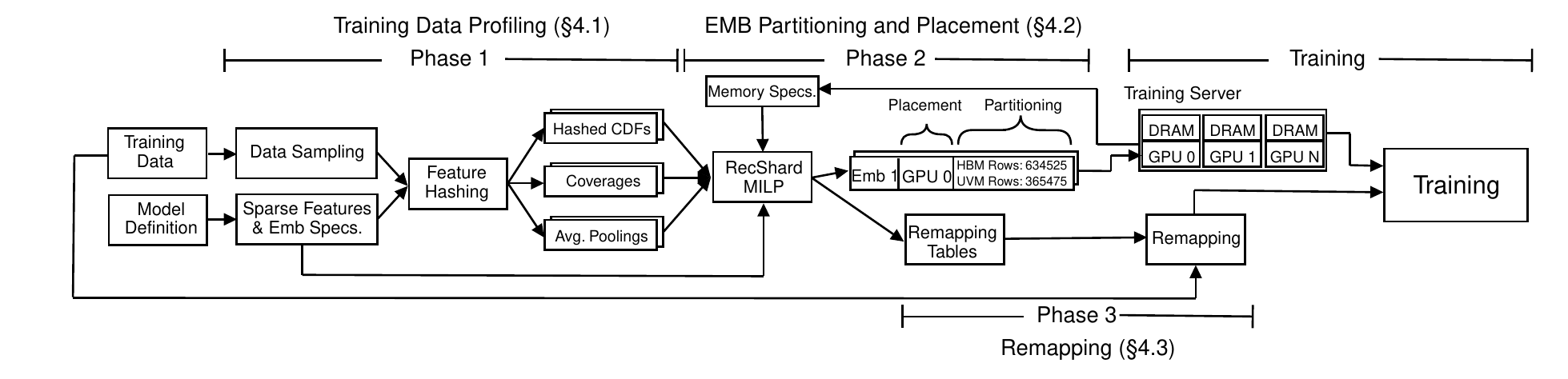}
    \caption{Overview flow diagram of the RecShard pipeline.}
  \label{fig:recshard}
\end{figure*}

Building on the EMB memory access characterization results in Section~\ref{sec:characterization}, we design, implement, and evaluate an intelligent EMB sharding strategy -- \emph{RecShard}.   
RecShard is a data-driven EMB sharding framework that optimizes embedding table partitioning and placement across a tiered memory hierarchy.
Figure~\ref{fig:recshard} provides the design overview for RecShard, which is comprised
of three primary phases: Training Data Profiling (Section~\ref{sec:profiling}), Embedding Table Partitioning and Placement (Section~\ref{sec:embedding-placement}), and Remapping (Section~\ref{sec:remapping}). RecShard leverages a MILP along with the latest training data distributions and EMB design characteristics to produce an optimal EMB sharding strategy \textit{each time a given DLRM is trained.}

\subsection{Training Data Profiling}
\label{sec:profiling}

The first stage of the RecShard pipeline is model-based training data profiling, which approximates the  aforementioned memory characteristics in Section~\ref{sec:characterization}. In this stage, RecShard first samples and hashes a random subset of the input training dataset based on the DLRM architecture specification. The purpose of this sampling is to estimate three per-EMB statistics: (1) the value frequency CDF over the EMB entries, (2) the average pooling factor of accesses for each EMB, and (3) each EMB's coverage over the training dataset.

We observe empirically that sampling 1\% or less of large training data stores achieves statistical significance to accurately facilitate high-performance EMB partitioning decisions. This is largely because increasing the sampling rate primarily serves to capture more of the tail of a sparse feature's skewed distribution. 
With respect to the value frequency CDF, these extra ``tail values,'' when hashed, will either map to their own EMB entry with minimal access count, or will collide with other previously-seen feature values. 
And viewing more of the tail has little to no impact on the average pooling factor and coverage of an EMB.
In all cases, not capturing the full tail is sufficient from the perspective of memory pattern profiling.

In the training data profiling phase, RecShard constructs the value frequency and pooling factor statistics as well as the coverage of each sparse feature for use in sharding.

\begin{table}[t]
\normalsize
\centering
\begin{tabular}{cl} 
\hline
\textbf{Parameter}                    & \textbf{Description}                      \\ 
\hline
\textit{M}                            & Number of GPUs                            \\
\textit{J}                            & Number of EMBs                            \\
\textit{B}     & Batch size       \\
\textit{Cap\textsubscript{D}}           & Per-GPU HBM Capacity                  \\
\textit{Cap\textsubscript{H}}           & Per-GPU Host DRAM Capacity                  \\
\textit{BW\textsubscript{HBM}}      & GPU HBM Bandwidth                         \\
\textit{BW\textsubscript{UVM}}      & UVM Transfer Bandwidth                    \\
\textit{ICDF\textsubscript{j}}       & Inverse Value Frequency CDF of EMB \textit{j}   \\
\textit{avg\_pool\textsubscript{j}} & Average Pooling Factor of EMB\textit{ j}  \\
\textit{coverage\textsubscript{j}}  & Coverage of EMB \textit{j}                \\
\textit{hash\_size\textsubscript{j}}  & Hash Size of EMB \textit{j}                \\
\textit{dim\textsubscript{j}}       & Embedding Dimension of EMB \textit{j}     \\
\textit{bytes\textsubscript{j}}     & Size of data-type of EMB \textit{j}       \\
\hline
\end{tabular}
\vspace{.15cm}
\caption{Description of Parameters used in the RecShard MILP.}
\label{table:milp_params}
\end{table}

\subsection{Embedding Table Partitioning and Placement}
\label{sec:embedding-placement} 

RecShard uses the generated per-feature statistics to produce an efficient, load-balanced EMB partitioning decision. In order to perform partitioning and sharding across multiple compute nodes with a tiered memory hierarchy, RecShard formulates the partitioning problem as a mixed integer linear program (MILP). By solving the MILP~\cite{gurobi}, RecShard can globally minimize per-GPU cost, a proxy for EMB training latency, simultaneously, while ensuring that neither GPU on-device nor per-node host memory limits are violated. The remainder of this section outlines our MILP formulation, which considers the problem of sharding EMBs across a two-tier memory hierarchy consisting of GPU HBM and host DRAM accessed via UVM. We refer to the latter as UVM for the rest of this paper. Table~\ref{table:milp_params} summarizes parameters used by the MILP formulation.

\noindent \textbf{MILP Formulation} \quad  As the training throughput is determined by the embedding operator performance of the slowest trainer, we formulate the MILP as a minimization problem to:
\begin{alignat}{3}
  \textit{minimize}   &            &&C  & \notag \\
\label{eq:obj}  \textit{subject to} & \quad\quad &&c_{m} \leq C \quad\quad & \forall m\in M
\end{alignat}
\noindent \textit{M} is the set of GPUs available for training (each GPU is represented by an integer ID $m$ ranging from 0 to $M - 1$), $c_{m}$ is the total EMB cost for GPU \textit{m}, and \textit{C} is the maximum single GPU cost to minimize, subject to Constraint~\ref{eq:obj}.

In order to estimate the total EMB cost per GPU, RecShard incrementally incorporates the per-EMB memory statistics to construct constraints which effectively describe the space of
all possible EMB partition and placement combinations for the underlying tiered memory hierarchy.

To construct a search space of candidate placements, 
the first constraint specified by RecShard is the mapping of each EMB to a single GPU. 
An EMB can either be located fully in a GPU's HBM, fully in UVM, or split across both in a fine-grained manner. If an EMB is placed entirely in HBM, the corresponding GPU will be the sole accessor of the entire EMB. 
If an EMB is placed entirely in UVM, it must be assigned a GPU that will issue memory accesses to it.
When an EMB is located in both HBM and UVM, we map both partitions to the GPU whose HBM is utilized. This constraint is formulated as follows: 
\begin{alignat}{2}
\label{eq:placement}  \sum_{m}p_{mj} = 1   &            &&\forall j\in J   \\
  p_{mj} \in \{0,1\} & \quad\quad &&\forall m\in M \quad \forall j\in J
\end{alignat}
\noindent $p_{mj}$ is a binary variable indicating whether EMB $j$ 
is assigned to GPU $m$, and Constraint \ref{eq:placement} ensures that each EMB is assigned to exactly one GPU.

When determining the EMB-to-GPU mappings, RecShard must also decide how many, or if any at all, of each EMB's rows should be placed in HBM. To do so, RecShard uses each EMB's post-hash value frequency CDF to estimate the trade-off between the number of rows placed in HBM and the corresponding percentage of EMB accesses covered.
To use the CDF within the MILP, RecShard first converts the CDF to its inverse, or ICDF, so that it can map the percentage of accesses covered to the corresponding number of EMBs rows. RecShard then produces a piece-wise linear approximation of the ICDF -- as the ICDF is a non-linear function, it cannot be used directly within the MILP.
To do so, 100 steps are uniformly selected with respect to the ICDF's $x$ values, where each step $i$ corresponds to a cumulative access percentage between 0 and 100\%. To capture both the $x$ and $y$ values of the ICDF, 
the constraints are formulated as follows:
\begin{alignat}{2}
\label{eq:mem_eq} \sum_{i}x_{ij} * ICDF_{j}(i) * dim_{j} * bytes_{j} = mem_{j}  &            &&\forall j\in J   \\
\label{eq:pct_eq} \sum_{i}x_{ij} * \frac{i}{100} = pct_{j} & \quad\quad &&\forall j\in J \\
\label{eq:split_eq} \sum_{i}x_{ij} = 1 & \quad\quad &&\forall j\in J \\
x_{ij} \in \{0,1\} \quad\quad i=0,...,100  & \quad\quad &&\forall j\in J 
\end{alignat}
\noindent $x_{ij}$ is a binary variable indicating whether step $i$ was chosen for EMB $j$. Constraint~\ref{eq:split_eq} ensures that one and only one step from the ICDF can be selected per EMB (i.e. there is a single split point separating the EMB rows mapped to HBM from those mapped to UVM). Constraint~\ref{eq:pct_eq} converts the chosen step value for each EMB into the corresponding percentage -- the ICDF's corresponding $x$ value. For each EMB, this percentage represents the cumulative percentage of accesses covered by the chosen split, and its value is stored as $pct_{j}$. Finally, Constraint~\ref{eq:mem_eq} translates each EMB's chosen split into the number of bytes needed to store its rows, $mem_{j}$ -- the per-EMB HBM usage. 

Given the constraints for encoding per-EMB HBM usage, 
constraints are added to guarantee per-GPU memory capacity limits are not violated.
\begin{alignat}{2}
\label{eq:emb_size} hash\_size_{j} * dim_{j} * bytes_{j} = EMB_{j}  &            &&\forall j\in J \\
\label{eq:mem_cap_hbm}  \sum_{j}p_{mj} * mem_{j} \leq Cap_{D}   & \quad\quad &&\forall m\in M \\
\label{eq:mem_cap_uvm}  \sum_{j}p_{mj} * (EMB_{j} - mem_{j}) \leq Cap_{H}   & \quad\quad &&\forall m\in M 
\end{alignat}
\noindent Constraint~\ref{eq:mem_cap_hbm} accomplishes this for per-GPU HBM by summing the memory capacity requirements of all EMB portions assigned to each GPU $m$ and ensuring that ensuring that no GPU exceeds its HBM capacity of $Cap_{D}$. 
Constraint~\ref{eq:mem_cap_uvm} accomplishes this similarly for per-GPU host DRAM capacity limits, $Cap_{H}$.

With the EMB partitioning and placement assignments properly constrained,
RecShard can formulate the
estimated per-GPU EMB cost.
\begin{alignat}{2}
(avg\_pool_{j} * dim_{j} * bytes_{j} * B) *   &            &&\notag   \\
((pct_{j}*\frac{1}{BW_{HBM}})+((1-pct_{j})*\frac{1}{BW_{UVM}}))   &            &&\notag   \\
\label{eq:emb_time} = c_{j}   &            &&\forall j\in J   \\
\sum_{j}p_{mj} * coverage_{j} * c_{j} = c_{m} & \quad   &&\forall m\in M
\label{eq:emb_time_coverage}
\end{alignat}
Constraint~\ref{eq:emb_time} estimates the cost of each EMB during a single forward pass of DLRM training. This is achieved by first calculating each EMB's approximate \textit{per-training step} memory demand using the EMB's average pooling factor, embedding vector dimension, size (in bytes) of its embedding vector entries, and 
batch size. Per-step demand is then multiplied by: (1) the percentage of EMB rows that are estimated to be sourced from HBM ($pct_{j}$) along with a bandwidth based scaling factor ($\frac{1}{BW_{HBM}}$); and (2) the percentage that are estimated to be sourced from UVM ($1-pct_{j}$) along with its scaling factor ($\frac{1}{BW_{UVM}}$). The two products are summed to form an estimate of an EMB's cost to perform a single step lookup on average.

Constraint~\ref{eq:emb_time_coverage} formulates the per-GPU cost for the MILP's objective function. Instead of simply summing the per-EMB costs assigned to a GPU, we sum the product of the per-EMB cost and the corresponding EMB's \emph{coverage}. This is because the per-EMB CDF presents a normalized view of accesses over \emph{a particular EMB}, and the average pooling factor estimates the EMB’s memory performance requirement \emph{over the samples it is present in}. Therefore, to provide a global view of bandwidth requirements \emph{across all EMBs}, RecShard weights each EMB’s cost by its coverage. 

With the constraints in place to formulate the per-GPU EMB cost, $c_{m}$, the MILP solver considers all possible combinations of EMB partitioning and placement decisions based on RecShard's EMB statistics and the bandwidth characteristics of the underlying memory hierarchy (supplied via the \textit{BW} parameters in Constraint~\ref{eq:emb_time}). In doing so, the MILP solver can compute an optimal sharding strategy that minimizes the model's largest single GPU EMB cost.

\noindent \textbf{Key Properties of RecShard's MILP} \quad
We address a number of key properties pertaining to RecShard's MILP formulation. 
First, when constructing its placements, RecShard's MILP begins by assigning each EMB to a single GPU. This design decision, and others which follow from it (such as per-GPU host DRAM capacity limits, $Cap_{H}$), is done to simplify the handling of sharding across many GPUs and nodes. By splitting resources uniformly and constructing assignments on an abstract per-GPU basis, the resulting sharding assignment will not be tied to a specific system GPU (i.e. GPU 3 in the MILP can be mapped to any GPU in the system).

Second, RecShard's MILP features \textit{summation} of the expected HBM and UVM lookup times to form a cost. Another operator, such as $max$, may be used, depending on the target system architecture. We use $summation$ in our implementation of RecShard because, 
when accessed within the same kernel, 
the memory latency of performing mixed reads from HBM and UVM on current GPUs is approximately equal to the time to perform each read separately. However, if the target system architecture supports concurrent reads fully from mixed memories, the estimated EMB cost can be approximated using $max$.

Third, while RecShard performs partitioning and placement for DLRM training, it only estimates embedding operation latencies of the forward pass in the MILP. This is because the timing performance of the backward pass is roughly proportional to its forward pass performance. By doing so, it simplifies the MILP formulation and lowers the solver time.

In our experiments in Section~\ref{sec:results}, RecShard's MILP features 47,276 variables, and is solved in 21 seconds when UVM is not needed (RM1), and in 42 seconds when used (RM2/RM3), with a state-of-the-art solver (Gurobi~\cite{gurobi}). It is important to note that solving time \textit{is not impacted} by model size, but instead in terms of the number of trainers (e.g. GPUs), and the steps used to approximate the ICDF. In our experiments solving time tended to scale approximately linearly with number of trainers and steps.

\subsection{Remapping Layer}
\label{sec:remapping}

Once the MILP solver produces a sharding strategy, RecShard determines the number of rows to be placed in HBM for each EMB via the activated $x_{ij}$ variable and the corresponding location on the EMB's ICDF, $ICDF_{j}(i)$.

These selected rows cannot be placed directly in HBM and must go through a remapping stage. This step is necessary as EMBs are typically allocated contiguously in memory, with an EMB index also serving as the memory offset to access the underlying storage directly. As the EMB rows selected by the MILP to be placed in HBM are chosen based on their access frequency, they can be located at arbitrary positions within the EMB and thus be non-contiguous. To address this, RecShard creates a per-EMB \emph{remapping table}, which maps each EMB index to its corresponding location in either HBM or UVM.

\subsection{Expansion Beyond Two-Tiers}

While the RecShard implementation is modeled after a two-tier memory hierarchy consisting of GPU HBM and host DRAM accessed via UVM, RecShard can be easily expanded to support a multi-tier memory hierarchy. At its core, each additional tier represents a new point on each EMB's CDF, potentially producing an additional split of EMB rows to be placed on the new memory tier. As each memory tier has its own bandwidth specifications from the view of the executing device (e.g. the GPU), the RecShard MILP solver will automatically order the memory tiers via the bandwidth scaling factors.

\section{Experimental Methodology}
\label{sec:methodology}

\noindent \textbf{Baselines:} \quad
To evaluate the efficacy of RecShard, we compare the performance of EMB operators under RecShard's throughput optimized sharding strategy with sharding schemes from prior work on production DLRM training systems~\cite{acun:2021:understandingtraining, lui:2021:capacity,mudigere2021highperformance}.
State-of-the art sharding schemes typically follow a two-step approach.
First, they assign a fixed cost to each EMB based on a specific cost function. Second, they apply a heuristic algorithm to incrementally assign EMBs to GPUs while attempting to minimize the maximum cost across all GPUs (a measure of load balancing).

\textbf{Step I--Cost Functions:} \quad
We implement the following three cost functions – two representing the state-of-the-art and a third derived from the first two – and compare their impact on EMB training throughput with RecShard:
\begin{itemize}\interlinepenalty10000
    \item {\bf Size}~\cite{acun:2021:understandingtraining, lui:2021:capacity}{\bf:}
    An EMB's cost is the product of its hash size and its embedding dimension (latent vector length).
    \item {\bf Lookup}~\cite{acun:2021:understandingtraining, lui:2021:capacity}{\bf:}
    An EMB's cost is the product of its average pooling factor and its embedding dimension.
    \item {\bf Size-and-Lookup:}
    An EMB's cost is the product of its lookup based cost (above) and 
    the log of its hash size -- \\ $log_{10}(hash\_size_{EMB})$ -- adding a non-linear function that attemps to capture potential caching effects of smaller EMBs.
\end{itemize}

\noindent In comparison, {\bf RecShard} considers EMB access distributions, average pooling factor, coverage, hash function, hash size, and the memory bandwidth characteristics of the target system.

\textbf{Step II--Heuristic Sharding Algorithms:} \quad
To shard EMBs once assigned a cost, we implement a \textbf{greedy heuristic algorithm}~\cite{mudigere2021highperformance} that works as follows. After receiving the list of EMBs to shard along with their associated costs, the greedy heuristic first sorts EMBs in descending cost order. It then descends the list, starting with the highest-cost EMB, and iteratively assigns EMBs to GPUs one-by-one. 
The heuristic continues down the sorted list of EMBs, placing each successive EMB on the GPU with the current lowest sum of costs. When GPU HBM has been saturated, the heuristic then allocates the remaining EMBs in UVM. 
In comparison, {\bf RecShard} considers cost on a per-\textit{EMB entry} basis and optimizes the placement of all EMB rows \textit{simultaneously, in one shot}.

\subsection{DLRM Specification}
\label{sec:specifications}
\begin{table}[]
\centering
\resizebox{\linewidth}{!}{%
\begin{tabular}{|c|c|c|c|c|}
\hline
Model & \# Sparse Features & Total Hash Size & Emb. Dim. & Size   \\ \hline
RM1   & 397    & 1,331,656,544  &      64      & 318 GB \\ \hline
RM2   & 397    & 2,661,369,917 &       64    & 635 GB \\ \hline
RM3   & 397    & 5,320,796,628 &     64      & 1270 GB \\ \hline
\end{tabular}}
\vspace{.15cm}
\caption{DLRM Specifications}
\label{table:model_desc}
\end{table}

We evaluate the performance of the different sharding strategies on a 
system running a modified version of open-source DLRM~\cite{dlrm_oss,naumov2019deep}. The implementation is modified to support the use of multi-hot encoded training data samples and the open-source implementation of the high-performance embedding operator in the PyTorch FBGEMM library\footnote{https://github.com/pytorch/FBGEMM/tree/master/fbgemm\_gpu}.

We implement the RecShard remapping layer as a custom PyTorch C++ operator which is executed as a transform during the data loading stage. This allows remapping to be performed in parallel with training iterations, thus removing it from the critical path of model execution.

We evaluate RecShard using three production-scale DLRMs: RM1, RM2, and RM3, summarized in Table~\ref{table:model_desc}.
All three RMs feature the same underlying DLRM architecture, implementing a large number of sparse features (397) and spanning a breadth of feature characteristics: categorical value distributions, pooling factors, and coverage;  which collectively determine the locality characteristics of embedding accesses (Section~\ref{sec:characterization}). The difference between the RMs is the approximate doubling of the hash size for each EMB from RM1 to RM2, and furthermore from RM2 to RM3.

We generate different workloads by having a large, constant number of features and scaling the hash sizes for two key reasons. First, the complexity of the sharding problem directly scales with the number of features to be sharded and their characteristics; thus, a large number of features maximizes sharding complexity. Second, as has been observed internally at our company, and in prior evaluations of industry-scale DLRMs~\cite{acun:2021:understandingtraining, kang2021learning,zhao:2020:distributedgpu}, increasing the hash size of an embedding table and thereby reducing collisions between sparse feature values is a simple, yet effective method of realizing accuracy improvements.

Based on the system specification discussed in the next section, RM1 requires 14 GPUs to completely fit all EMBs in reserved HBM, while RM2 requires 27 GPUs, and RM3 requires 53 GPUs.

\subsection{Training System Specification}
We evaluate the timing performance for all three sharding strategies 
on a two-socket server-node.
Each socket features an Intel Xeon Platinum 8339HC CPU, 376GB of DDR4 DRAM, and 8x NVIDIA A100 (40GB) GPUs, connected to host DRAM via PCIe 3.0x16 for UVM support. As the scale of the RMs exceeds that of the memory capacity of the training nodes, 
during benchmarking we run each model-parallel section separately and extract the EMB performance metrics.

When implementing the training sharding strategies from prior work~\cite{acun:2021:understandingtraining, lui:2021:capacity} (our baselines for comparison), we use a batch size of 16,384 and limit each sharding strategy to use at most: (1) 24GB of HBM per GPU as the reserved memory for EMBs; (2) 128GB of host DRAM for usage per GPU for UVM allocated EMBs. The remaining HBM/DRAM capacity reserved for other model parameters, computation, and training overheads.

\textbf{Performance Profiling:} \quad
As the goal of RecShard is to improve per-iteration EMB latency, due to the large percentage of total run-times they constitute for many types of DLRMs~\cite{acun:2021:understandingtraining, gupta:2020:archimp, zhao:2020:distributedgpu}, we evaluate execution time by tracing each GPUs execution and extracting all kernel run times related to the embedding operator. We do this by using the integrated PyTorch profiler,  \verb|torch.profiler|,
which allows for tracing to begin after a specified waiting and warm-up period. We specify a waiting period of 10 iterations, a warm-up period of 5 iterations, and trace for 20 iterations. 

\section{Evaluation Results and Analysis}
\label{sec:results}

\begin{table*}
\centering
\begin{tabular}{c|c|c|c|c}
  Model  & Size-Based & Lookup-Based & Size-Based-Lookup & RecShard  \\ 
\hline
RM1 &   7.12/21.23/13.06/4.01   &   5.08/30.97/12.99/5.59     &      5.55/26.03/12.91/4.72         &     6.53/\textbf{8.21}/7.48/\textbf{0.45}      \\ 
\hline
RM2 &   20.52/49.65/33.82/7.37   &   10.40/55.85/32.47/9.87     &      7.47/56.66/32.95/10.26         &    6.52/\textbf{9.44}/7.75/\textbf{0.78}       \\ 
\hline
RM3 &   40.43/76.15/56.45/10.86   &   3.37/73.30/55.27/18.53     &       5.10/85.01/56.04/20.39        &      6.83/\textbf{9.90}/8.31/\textbf{0.69}     \\
\hline
\end{tabular}
\vspace{.15cm}
\caption{Min/Max/Mean/StdDev EMB training iteration time (in ms) across all GPUs based on per GPU averages for all sharding strategies on 16 GPUs. Training performance is bound by the slowest (i.e. \textit{max}) EMB time, therefore lowest max iteration time is better. Load balanced is embodied by the standard deviation, with lower standard deviation signifying more balanced execution.}
\label{table:results}
\end{table*}

\begin{figure}[t]
  \centering
    \includegraphics[width=.90\linewidth,keepaspectratio]{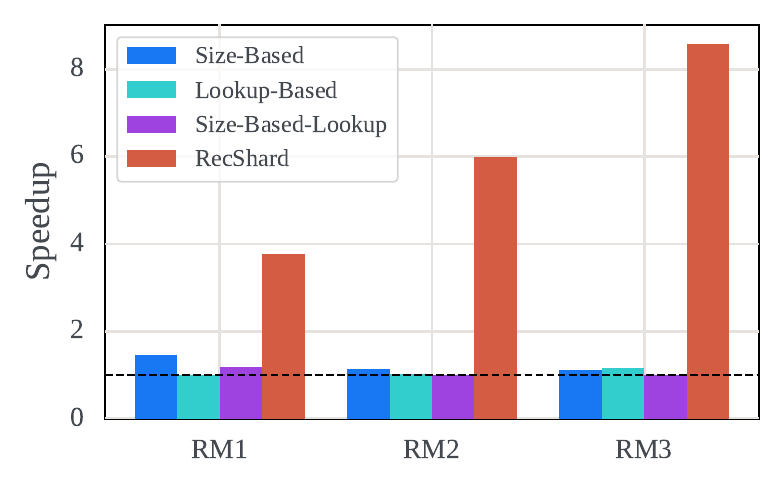}
    \caption{EMB training performance improvement of different sharding strategies normalized to slowest strategy in group. RM1, RM2, and RM3 evaluated using 16 GPUs.}
  \label{fig:perf_comp}
\end{figure}

Overall, RecShard achieves an average of 5x EMB training iteration time improvement across RM1, RM2, and RM3 in the 16-GPU setting, covering a wide range of memory demands.
Figure~\ref{fig:perf_comp} illustrates that RecShard improves the EMB training iteration time by 2.58x, 5.26x, and 7.41x for RM1, RM2, and RM3, respectively, over the next fastest sharding strategy. 
Table~\ref{table:results} summarizes the timing results for RecShard and the state-of-the-art schemes.

\subsection{RecShard Workload Balance Analysis}

A major factor contributing to RecShard's significant performance improvement comes from its ability to achieve better load balance across the GPU trainers.
In particular, the EMB memory footprint of RM1 is approximately 318GB, allowing all EMBs to fit entirely in HBM when distributed among the 16 available GPUs.
RecShard improves EMB training throughput
for RM1 by over 2.5x with respect to the next fastest sharding strategy  (\textbf{Size}). It does so with an almost 9 times improvement in the standard-deviation of the iteration time across all GPUs, providing a much more uniform distribution of work (Table~\ref{table:results}).

RecShard's ability to better load balance comes from two key aspects of its design. First, RecShard's hash analysis allows it to effectively determine \textit{which} portion of each EMB is unused or sparsely used during training. The sparse regions are
effectively assigned a cost of zero
and thus can be allocated last. Second, RecShard's formulation of the EMB sharding problem as a MILP allows it to globally balance EMB operations across all GPUs \textit{simultaneously, in one shot}. Since RM1 does not require UVM for EMB placement, the sharding cost formulation reduces to a function that is similar to the \textbf{Lookup} cost function of Section~\ref{sec:specifications}. However, when used with the greedy heuristic, the \textbf{Lookup} sharding strategy performs 46\% worse than the \textbf{Size} strategy (the best baseline RM1 strategy). This result highlights the performance improvements that stem from RecShard's fine-grained, data-driven MILP approach to embedding vector placement. 

\subsection{RecShard Embedding Placement Analysis}

As DLRM sizes grow beyond the capacity of available GPU HBM, as is the case for RM2 and RM3, sharding pressure moves beyond simply load balancing across HBM and into load balancing across HBM and UVM. With orders of magnitude difference in the memory performance of HBM and UVM, incorrect EMB placements on UVM come with severe performance penalties. In this scenario, the state-of-the-art sharding strategies can significantly under-perform RecShard. This is exemplified with RM2's and RM3's results.

RecShard uses feature and EMB statistics to dynamically estimate EMB cost at the row granularity, enabling it to intelligently break apart an EMB into non-contiguous memory blocks and place each block independently across different tiers of the memory hierarchy. By doing so, RecShard determines and places the least performance-critical embedding portions of large DLRMs (i.e. RM2 and RM3) onto UVM. 

\begin{figure}[t]
  \centering
    \includegraphics[width=.90\linewidth,keepaspectratio]{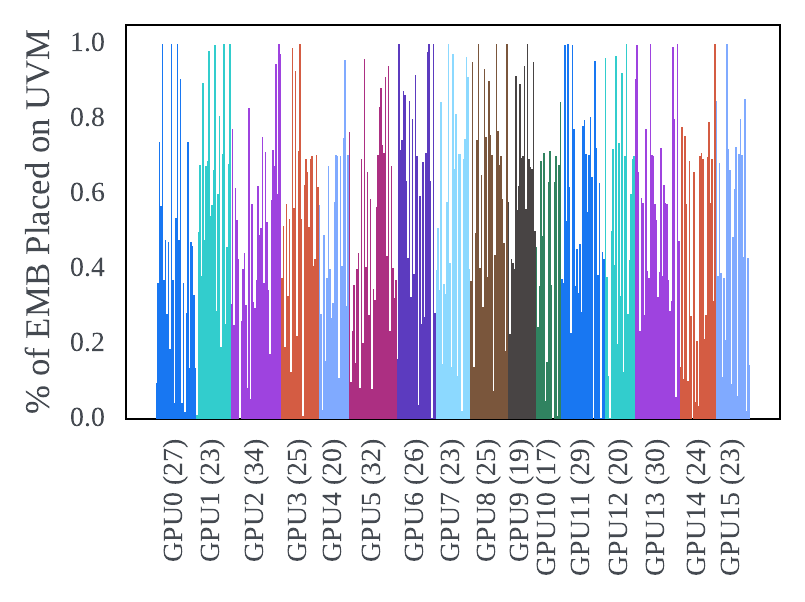}
    \caption{Partitions and Placements made by RecShard for RM2 on 16 GPUs. Each bar represents a single EMB. Bar height is the percentage of a specific EMB that RecShard placed on UVM. EMBs are grouped by the GPU they were assigned to (shown as colors). The number of EMBs assigned to each GPU is shown in parentheses. As expected, the number of EMBs assigned to each GPU is variable and the height of each bar is unique to each EMB.}
  \label{fig:emb_placements}
\end{figure}

For RM2, RecShard places an average of 53.4\% of rows per EMB and a total of 61\% of all EMB rows on UVM. For RM3, it places an average of 64.8\% of rows per EMB and a total of 61\% of all EMB rows on UVM. Figure~\ref{fig:emb_placements} illustrates the partitioning decisions for RM2 using RecShard.

\begin{table}
\centering
\resizebox{\linewidth}{!}{%
\begin{tabular}{c|c|c|c|c}
                Model               &    Disparity     & \multicolumn{1}{c|}{SB} & \multicolumn{1}{c|}{LB} & \multicolumn{1}{c}{SBL}  \\ 
\hline
\multirow{2}{*}{RM1} & UVM->HBM & N/A                         & N/A                           & N/A                                 \\ 
\cline{2-5}
                               & HBM->UVM & N/A                         & N/A                           & N/A                                 \\
\hline
\multirow{2}{*}{RM2} & UVM->HBM & 28.67\%                         & 28.26\%                           & 28.26\%                                 \\ 
\cline{2-5}
                               & HBM->UVM & 39.93\%                         & 39.99\%                           & 39.99\%                                 \\ 
\hline
\multirow{2}{*}{RM3} & UVM->HBM & 23.29\%                         & 23.21\%                           & 23.21\%                                 \\ 
\cline{2-5}
                               & HBM->UVM & 58.34\%                         & 59.36\%                           & 59.36\%                                 \\
\hline
\end{tabular}}
\vspace{.15cm}
\caption{Percent of EMB rows allocated in UVM (resp. HBM) by each baseline strategy which RecShard places in HBM (resp. UVM).
RM2 and RM3 require UVM for training on 16 GPUs, whereas RM1 does not.
LB and SBL assign the same EMBs to HBM and UVM, but their exact GPU assignments differ.
SB, LB, and SBL stand for Size-Based, Lookup-Based, and Size-Based-Lookup, respectively.}
\label{table:emb_placements1}
\end{table}

\begin{table}
\centering
\resizebox{\linewidth}{!}{%
\begin{tabular}{c|c|c|c|c|c}
\multicolumn{1}{c|}{Model}     & Location & SB   & LB   & SBL  & RecShard  \\ 
\hline
\multirow{2}{*}{RM1} & HBM      & 88.74M & 88.74M & 88.74M & 88.74M      \\ 
\cline{2-6}
                               & UVM      & 0 & 0 & 0 & 0         \\ 
\hline
\multirow{2}{*}{RM2} & HBM      & 70.32M & 70.90M & 70.90M & 88.48M      \\ 
\cline{2-6}
                               & UVM      & 18.42M & 17.84M & 17.84M & 259K         \\ 
\hline
\multirow{2}{*}{RM3} & HBM      & 55.82M & 56.85M & 56.85M & 88.29M      \\ 
\cline{2-6}
                               & UVM      & 32.92M & 31.89M & 31.89M & 450K         \\
\hline
\end{tabular}}
\vspace{.15cm}
\caption{Average number of HBM and UVM accesses per-GPU, per-iteration for each sharding strategy on RM2 and RM3 (batch-size of 16384 on 16 GPUs). RM1 does not not require UVM. Baseline strategies source on average 20.3\% (RM2) and 36.3\% (RM3) of EMB accesses from UVM. RecShard sources 0.2\% (RM2) and 0.5\% (RM3) of EMB accesses from UVM. LB and SBL assign the same EMBs to HBM and UVM, but their exact GPU assignments differ. SB, LB, and SBL stand for Size-Based, Lookup-Based, and Size-Based-Lookup, respectively.}
\label{table:emb_placements2}
\end{table}

To further understand the difference in decision making between the baseline strategies and RecShard, we compare the EMB assignments and expected access counts for all strategies across RM2 and RM3. 
First, we explore how the individual EMB assignment by the \textbf{Size}, \textbf{Lookup}, and \textbf{Size-and-Lookup} strategies differ from RecShard's placement. That is, if an EMB was assigned to HBM, we examine the degree of overlap for the rows placed on UVM between the state-of-the-art strategy and RecShard.
Table~\ref{table:emb_placements1} summarizes this analysis. The rows labeled `UVM->HBM' quantify the difference in the percentage of EMB rows placed in UVM for RM2 and RM3 by the state-of-the-art strategies versus RecShard. RecShard's ability to place more performance-critical, frequently-accessed embedding vectors onto HBM across all EMBs is the primary reason for its significantly higher performance.

\subsection{RecShard Scalability Analysis}
\label{sec:scalability-analysis} 

\begin{figure}[t]
  \centering
    \includegraphics[width=.90\linewidth,keepaspectratio]{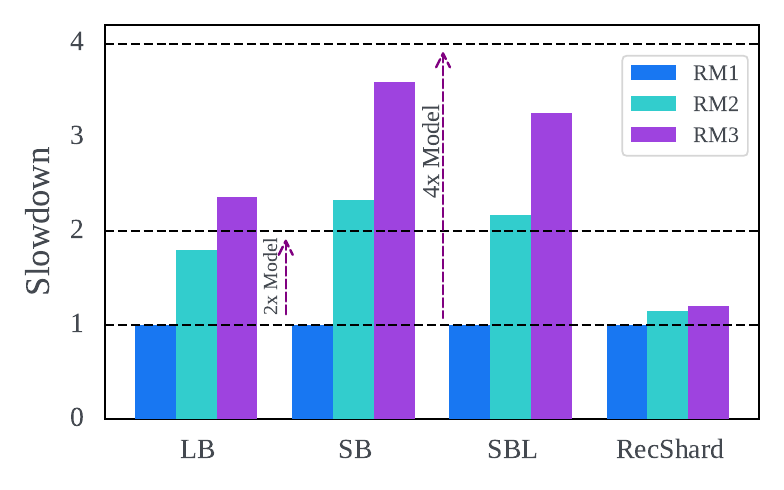}
    \caption{Slowdown of each sharding strategy on max EMB iteration time as model sizes scale 2x and 4x from RM1 to RM2, and RM1 to RM3, respectively. While heuristic based fixed-cost strategies suffer over a 3x slowdown on average from RM1 to RM3, RecShard is less sensitive to performance degradation from model size scaling and only experiences a 1.2x slowdown. 
    }
  \label{fig:slowdowns}
\end{figure}

As model sizes increase, as expected, RecShard sees little performance degradation. This comes from the asymmetric impact on memory access statistics and memory usage that hash size scaling causes. The state-of-the-art strategies experience an average of 3.07 times performance slowdown in the EMB training iteration time between the largest DLRM (RM3) and RM1. However, RecShard only observes a 20.6\% increase in the EMB training iteration time over the same model size growth (Figure~\ref{fig:slowdowns}).

We explored this sparsity in access count by analysing the number of HBM and UVM accesses made by the EMBs in each of the sharding strategies in our training traces. We found (Table~\ref{table:emb_placements2}) that when doubling the hash size from RM1 to RM2, the baseline sharding strategies sourced on average 20.3\% of their accesses per-GPU per-iteration from UVM, while RecShard only sourced 0.2\% -- \textit{over a 100x reduction}. When hash size is quadrupled from RM1 to RM3 and sharding pressure doubles from RM2, the baseline sharding strategies sourced on average 36.3\% of their accesses per-GPU per-iteration from UVM, while RecShard only sourced 0.5\%. As HBM capacity is already exceeded in RM2, the additional model capacity (in bytes) of RM3 \textit{must} be allocated in UVM. While the percentage of accesses sourced from UVM for RecShard more than doubles from RM2 to RM3, this value is still only 0.5\% of the total accesses (and \textit{over 70x less} than the baseline strategies). This result highlights the sparsity of memory access to the new memory regions allocated by increased hash size.

\subsection{End-to-End Training Time Improvement}

While embedding operations can represent a significant portion of many industry-scale DLRMs~\cite{gupta:2020:archimp, zhao:2020:distributedgpu}, the actual percentage of runtime varies based on model composition. RecShard improves end-to-end training performance in proportion to the time spent on embedding operations in the critical path of model execution (which in the canonical DLRM architecture consists of all embedding operations).

Knowing the runtime breakdown, the expected end-to-end DLRM training performance improvement can be 
approximated using Amdahl’s law.
With $p$ being the percentage of total execution time spent on critical path embedding operations, and $s$ being the speedup in embedding operation latency via improved sharding, the estimated end-to-end speedup is $\frac{1}{(1-p)+\frac{p}{s}}$.

As a concrete example, for memory-intensive models whose timing composition consists of 35-75\% embedding operations~\cite{gupta:2020:archimp, rec-nmp} (with the remaining time being largely dominated by dense MLP layers and communication), and for which RecShard improves embedding performance by 2.5x, the expected end-to-end performance benefit of RecShard is 1.27x to 1.82x. While the performance improvements afforded by RecShard are less pronounced for more MLP-dominated DLRMs, the position of embedding operations on the critical path of model execution and the scale of industry-DLRM training time (on the order of days~\cite{acun:2021:understandingtraining}) indicates the importance of their acceleration. 

\subsection{RecShard Ablation}
\label{sec:recshard-ablation}

\newcolumntype{Y}{>{\centering\arraybackslash}X}
\begin{table}
\centering
\begin{tabularx}{\linewidth}{X|Y|Y}

Formulation & HBM & UVM     \\
\hline
RecShard (Full) & 69.07B & 353M \\
\hline
CDF + Pooling & 68.82B & 604M \\
\hline
CDF + Coverage & 68.54B & 881M \\
\hline
CDF Only & 67.79B & 1.63B \\
\hline
\end{tabularx}
\vspace{.15cm}
\caption{RecShard Ablation. Average number of HBM and UVM accesses per-GPU on RM3 (across 16 GPUs) over more than 200 million training data samples for different RecShard formulations. \textit{CDF only} is the use of only the per-sparse feature value CDF in the MILP (i.e. average pooling factor and coverage are set 1). \textit{CDF + Coverage} is the use of both the CDF and coverage in the MILP; while \textit{CDF + Pooling} is the use of both the CDF and average pooling factor in the MILP. \textit{RecShard (Full)} is the access counts when all per-EMB statistics are used simultaneously in the MILP.}
\label{table:ablation}
\end{table}

To better understand the impact the various sparse feature characteristics used within RecShard have on the performance of the generated sharding, we performed an ablation study to measure their significance on the number of HBM and UVM accesses made by each GPU.
We evaluate four different formulation of RecShard, each differing by which per-EMB statistics are enabled for use within the MILP. The results of this ablation on RM3 (with 16 GPUs) over more than 200 million training data samples is shown in Table~\ref{table:ablation}. 
The four formulations of RecShard evaluated are: 
\begin{itemize}
\item \textit{CDF only}: Only the sparse feature value CDF is used in the MILP and the average pooling factor and coverage for each EMB are set to 1.
\item \textit{CDF + Coverage}: Both the CDF and the per-EMB coverage are used in the MILP.
\item \textit{CDF + Pooling}: Both the CDF and the per-EMB average pooling factor are used in the MILP. 
\item \textit{Full}: All of the per-EMB statistics are used in the MILP simultaneously. 
\end{itemize}

Similar to the results in Section~\ref{sec:scalability-analysis}, we observe that approximately 0.5\% of accesses on average in the full formulation of RecShard are sourced from UVM, while the simplest RecShard formulation, \textit{CDF only}, sources approximately 2.4\% of its accesses from UVM. While this is still significantly less than the baseline sharding strategies, this nearly 5x increase over the full formulation is due to the CDF providing no information about \textit{how often} each EMB will be accessed in a training data sample.
Thus when evaluating different potential partitioning and placement decisions, the MILP in the \textit{CDF only} formulation has no information which it can exploit to accurately load balance EMBs across the GPUs based on their expected number of accesses. 
Adding one piece of per-sample EMB access information via the coverage almost halves the average UVM sourced access percentage to approximately 1.3\%, while using the average pooling factor instead provides an even greater reduction to approximately 0.9\%. 

\subsection{RecShard Overhead}
\label{sec:recshard-overhead}

For all models studied in this work, the Gurobi solver~\cite{gurobi} was able to solve the placement and partitioning MILP in under 1 minute. After which, generating the remapping tables takes approximately 20 seconds for each GPU and has a storage cost of 4 bytes per row remapped (as the sign of the remapped index can be used to denote whether the corresponding table is the HBM or UVM partition). For the largest DLRM--RM3, this is a total storage overhead of \char`\~20GB for over 5-billion rows remapped. In the scope of model training time (many hours to potentially days depending on model and data size), and model size (hundreds of GBs to multiple TBs), this overhead is minimal, especially due to the performance improvements RecShard provides.  

Additionally RecShard incurs some overhead from training data profiling due to the consumption of feature level statistics. However, besides only needing to sample a small portion (\char`\~1\%) of large training data stores to achieve statistical significance, as the statistics are based on raw training data values and corresponding hash sizes (which are generally constant across models within a size tier), they can be shared across models and also updated in real-time as training data arrives, amortizing the cost.

\section{Related Work}
\label{sec:related}

Power-law distributions are a well-known phenomenon of features related to recommender systems~\cite{adnan:2021:hotsplit, ginart2019mixed, kang2021learning, wu:2020:mlperfrec, freq_hashing}. This sparsity characteristic is an important feature for a variety of DL system performance optimizations. 
However, maintaining the long tail is important because of the statistically significant accuracy impact~\cite{zhao:2020:distributedgpu}.
This has led to recent works attempting to balance the trade-off between EMB sizes and model accuracy. One such category of work explores scaling the dimension of an EMB, that is the number of parameters used to encode an EMB row, based on the frequency of accesses to individual rows---more frequently accessed rows are given more space through increased embedding vector dimensions~\cite{ginart2019mixed}. Another work explores the impact of hashing, ranging from the use of multiple hash functions alongside a 1:1 mapping for frequent categorical values~\cite{freq_hashing}, to entirely replacing the hashing plus embedding table structure with its own neural network~\cite{kang2021learning}. In addition, other prior work proposes to prioritize frequently-accessed embedding rows for model parameter checkpointing~\cite{maeng:2021:cpr}, in order to improve failure tolerance of DLRM training. While prior work also tackles the problem of ever-increasing EMB sizes, their primary focus is the size of EMB itself, rather than on training throughput improvement. 

Recent work has also explored the performance of splitting EMBs based on their frequency characteristics~\cite{adnan:2021:hotsplit}. While similar in motivation,
the type of training data and the scale of DLRMs explored in this paper  
are fundamentally different from the open-source datasets used in the related work. Our DLRMs read \textit{multi-hot encoded sparse features} resulting in order-of-magnitude higher memory bandwidth needs, and EMB sizes demanding model-parallel training. 
In Criteo Terabyte (the largest of the open-source datasets), all of the features are 1-hot encoded (meaning their pooling factor is always 1), the number of features present is 26, and the total number of un-hashed embedding table rows is approximately 266 million. Thus, for each of these properties, the scale of open-source datasets/DLRMs~\cite{mattson:2020:mlperftraining,reddi:2020:mlpefinference,wu:2020:mlperfrec} is an order of magnitude (or more) less than our evaluated datasets/DLRMs. Furthermore, all open-source datasets that we are aware of can fit entirely within a single GPU, making sharding and model-parallel training unnecessary.

\section{Conclusion}
\label{sec:conclusion}

Deep learning recommendation systems are the backbone of a wide variety of cloud services and products. 
Unlike other neural networks with primarily convolution or fully-connected layers, recommendation model embedding tables demand orders-of-magnitude higher memory capacity (>99\% of the model capacity) and bandwidth, and exhibit significantly lower compute-intensity. In this paper, we perform an in-depth memory characterization analysis and  we identify five important memory characteristics for sparse features of DLRMs.
Building on the analysis, we propose RecShard, which formulates the embedding table partitioning and placement problem for training systems with tiered memories. RecShard uses a MILP to reach a partitioning and placement decision that minimizes embedding access time under constrained memory capacities. We implement and evaluate RecShard by training a modified version of open-source DLRM with production data. RecShard can achieve an average of over 5 times speedup for the embedding kernels of three representative industry-scale recommendation models. We hope our findings will lead to further memory optimization insights in this important category of deep learning use cases.

\section*{ACKNOWLEDGMENTS}
We would like to thank Jade Nie, Jianyu Huang, Jongsoo Park, Andrew Tulloch, Xing Liu, Benny Chen, Ying Liu, Liu Ke, Udit Gupta, Newsha Ardalani, Hsien-Hsin S. Lee, and Kim Hazelwood at Meta for their valuable feedback and various discussions on this work, as well as Fan Yang and the anonymous reviewers for their constructive feedback. This work was supported in part by the Stanford Platform Lab and its affiliates for Geet Sethi and Christos Kozyrakis.

\bibliographystyle{ACM-Reference-Format}
\balance
\bibliography{references}


\begin{thebibliography}{48}

%%% the Web address.

\ifx \showCODEN    \undefined \def \showCODEN     #1{\unskip}     \fi
\ifx \showDOI      \undefined \def \showDOI       #1{#1}\fi
\ifx \showISBNx    \undefined \def \showISBNx     #1{\unskip}     \fi
\ifx \showISBNxiii \undefined \def \showISBNxiii  #1{\unskip}     \fi
\ifx \showISSN     \undefined \def \showISSN      #1{\unskip}     \fi
\ifx \showLCCN     \undefined \def \showLCCN      #1{\unskip}     \fi
\ifx \shownote     \undefined \def \shownote      #1{#1}          \fi
\ifx \showarticletitle \undefined \def \showarticletitle #1{#1}   \fi
\ifx \showURL      \undefined \def \showURL       {\relax}        \fi
% invisible to TeX
\providecommand\bibfield[2]{#2}
\providecommand\bibinfo[2]{#2}
\providecommand\natexlab[1]{#1}
\providecommand\showeprint[2][]{arXiv:#2}

\bibitem[\protect\citeauthoryear{Acun, Murphy, Wang, Nie, Wu, and
  Hazelwood}{Acun et~al\mbox{.}}{2021}]%
        {acun:2021:understandingtraining}
\bibfield{author}{\bibinfo{person}{Bilge Acun}, \bibinfo{person}{Matthew
  Murphy}, \bibinfo{person}{Xiaodong Wang}, \bibinfo{person}{Jade Nie},
  \bibinfo{person}{Carole-Jean Wu}, {and} \bibinfo{person}{Kim Hazelwood}.}
  \bibinfo{year}{2021}\natexlab{}.
\newblock \showarticletitle{Understanding Training Efficiency of Deep Learning
  Recommendation Models at Scale}. In \bibinfo{booktitle}{\emph{2021 IEEE
  International Symposium on High Performance Computer Architecture (HPCA)}}.
\newblock


\bibitem[\protect\citeauthoryear{Adnan, Maboud, Mahajan, and Nair}{Adnan
  et~al\mbox{.}}{2021}]%
        {adnan:2021:hotsplit}
\bibfield{author}{\bibinfo{person}{Muhammad Adnan},
  \bibinfo{person}{Yassaman~Ebrahimzadeh Maboud}, \bibinfo{person}{Divya
  Mahajan}, {and} \bibinfo{person}{Prashant~J. Nair}.}
  \bibinfo{year}{2021}\natexlab{}.
\newblock \showarticletitle{High-Performance Training by Exploiting
  Hot-Embeddings in Recommendation Systems}.
\newblock \bibinfo{journal}{\emph{CoRR}} (\bibinfo{year}{2021}).
\newblock
\urldef\tempurl%
\url{https://arxiv.org/abs/2103.00686}
\showURL{%
\tempurl}


\bibitem[\protect\citeauthoryear{Cheng, Koc, Harmsen, Shaked, Chandra, Aradhye,
  Anderson, Corrado, Chai, Ispir, Anil, Haque, Hong, Jain, Liu, and Shah}{Cheng
  et~al\mbox{.}}{2016}]%
        {Cheng:dlrs2016}
\bibfield{author}{\bibinfo{person}{Heng-Tze Cheng}, \bibinfo{person}{Levent
  Koc}, \bibinfo{person}{Jeremiah Harmsen}, \bibinfo{person}{Tal Shaked},
  \bibinfo{person}{Tushar Chandra}, \bibinfo{person}{Hrishi Aradhye},
  \bibinfo{person}{Glen Anderson}, \bibinfo{person}{Greg Corrado},
  \bibinfo{person}{Wei Chai}, \bibinfo{person}{Mustafa Ispir},
  \bibinfo{person}{Rohan Anil}, \bibinfo{person}{Zakaria Haque},
  \bibinfo{person}{Lichan Hong}, \bibinfo{person}{Vihan Jain},
  \bibinfo{person}{Xiaobing Liu}, {and} \bibinfo{person}{Hemal Shah}.}
  \bibinfo{year}{2016}\natexlab{}.
\newblock \showarticletitle{Wide \& Deep Learning for Recommender Systems}. In
  \bibinfo{booktitle}{\emph{Workshop on Deep Learning for Recommender
  Systems}}.
\newblock


\bibitem[\protect\citeauthoryear{Covington, Adams, and Sargin}{Covington
  et~al\mbox{.}}{2016}]%
        {covington:2016:youtuberec}
\bibfield{author}{\bibinfo{person}{Paul Covington}, \bibinfo{person}{Jay
  Adams}, {and} \bibinfo{person}{Emre Sargin}.}
  \bibinfo{year}{2016}\natexlab{}.
\newblock \showarticletitle{Deep Neural Networks for {YouTube}
  Recommendations}. In \bibinfo{booktitle}{\emph{ACM Recommender Systems
  Conference}}.
\newblock


\bibitem[\protect\citeauthoryear{Deng, Dong, Socher, Li, Li, and Fei-Fei}{Deng
  et~al\mbox{.}}{2009}]%
        {imagenet}
\bibfield{author}{\bibinfo{person}{Jia Deng}, \bibinfo{person}{Wei Dong},
  \bibinfo{person}{Richard Socher}, \bibinfo{person}{Li-Jia Li},
  \bibinfo{person}{Kai Li}, {and} \bibinfo{person}{Li Fei-Fei}.}
  \bibinfo{year}{2009}\natexlab{}.
\newblock \showarticletitle{ImageNet: A large-scale hierarchical image
  database}. In \bibinfo{booktitle}{\emph{2009 IEEE Conference on Computer
  Vision and Pattern Recognition}}. \bibinfo{pages}{248--255}.
\newblock
\urldef\tempurl%
\url{https://doi.org/10.1109/CVPR.2009.5206848}
\showDOI{\tempurl}


\bibitem[\protect\citeauthoryear{Devlin, Chang, Lee, and Toutanova}{Devlin
  et~al\mbox{.}}{2019}]%
        {devlin-etal-2019-bert}
\bibfield{author}{\bibinfo{person}{Jacob Devlin}, \bibinfo{person}{Ming-Wei
  Chang}, \bibinfo{person}{Kenton Lee}, {and} \bibinfo{person}{Kristina
  Toutanova}.} \bibinfo{year}{2019}\natexlab{}.
\newblock \showarticletitle{{BERT}: Pre-training of Deep Bidirectional
  Transformers for Language Understanding}. In
  \bibinfo{booktitle}{\emph{Proceedings of the 2019 Conference of the North
  {A}merican Chapter of the Association for Computational Linguistics: Human
  Language Technologies, Volume 1 (Long and Short Papers)}}.
  \bibinfo{publisher}{Association for Computational Linguistics},
  \bibinfo{address}{Minneapolis, Minnesota}, \bibinfo{pages}{4171--4186}.
\newblock
\urldef\tempurl%
\url{https://doi.org/10.18653/v1/N19-1423}
\showDOI{\tempurl}


\bibitem[\protect\citeauthoryear{{Facebook Research}}{{Facebook
  Research}}{2021}]%
        {dlrm_oss}
\bibfield{author}{\bibinfo{person}{{Facebook Research}}.}
  \bibinfo{year}{2021}\natexlab{}.
\newblock \bibinfo{title}{An implementation of a deep learning recommendation
  model (DLRM)}.
\newblock
\newblock
\newblock
\shownote{\url{https://github.com/facebookresearch/dlrm}}.


\bibitem[\protect\citeauthoryear{Ginart, Naumov, Mudigere, Yang, and
  Zou}{Ginart et~al\mbox{.}}{2019}]%
        {ginart2019mixed}
\bibfield{author}{\bibinfo{person}{Antonio Ginart}, \bibinfo{person}{Maxim
  Naumov}, \bibinfo{person}{Dheevatsa Mudigere}, \bibinfo{person}{Jiyan Yang},
  {and} \bibinfo{person}{James Zou}.} \bibinfo{year}{2019}\natexlab{}.
\newblock \showarticletitle{Mixed Dimension Embeddings with Application to
  Memory-Efficient Recommendation Systems}.
\newblock \bibinfo{journal}{\emph{arXiv preprint arXiv:1909.11810}}
  (\bibinfo{year}{2019}).
\newblock


\bibitem[\protect\citeauthoryear{Gomez-Uribe and Hunt}{Gomez-Uribe and
  Hunt}{2016}]%
        {netflix}
\bibfield{author}{\bibinfo{person}{Carlos~A. Gomez-Uribe} {and}
  \bibinfo{person}{Neil Hunt}.} \bibinfo{year}{2016}\natexlab{}.
\newblock \showarticletitle{The Netflix Recommender System: Algorithms,
  Business Value, and Innovation}.
\newblock \bibinfo{journal}{\emph{ACM Trans. Manage. Inf. Syst.}}
  \bibinfo{volume}{6}, \bibinfo{number}{4}, Article \bibinfo{articleno}{13}
  (\bibinfo{date}{Dec.} \bibinfo{year}{2016}), \bibinfo{numpages}{19}~pages.
\newblock
\showISSN{2158-656X}
\urldef\tempurl%
\url{https://doi.org/10.1145/2843948}
\showDOI{\tempurl}


\bibitem[\protect\citeauthoryear{Gupta, Hsia, Saraph, Wang, Reagen, Wei, Lee,
  Brooks, and Wu}{Gupta et~al\mbox{.}}{2020a}]%
        {deeprecsys}
\bibfield{author}{\bibinfo{person}{Udit Gupta}, \bibinfo{person}{Samuel Hsia},
  \bibinfo{person}{Vikram Saraph}, \bibinfo{person}{Xiaodong Wang},
  \bibinfo{person}{Brandon Reagen}, \bibinfo{person}{Gu-Yeon Wei},
  \bibinfo{person}{Hsien-Hsin~S. Lee}, \bibinfo{person}{David Brooks}, {and}
  \bibinfo{person}{Carole-Jean Wu}.} \bibinfo{year}{2020}\natexlab{a}.
\newblock \showarticletitle{{DeepRecSys: A} System for Optimizing End-To-End
  At-Scale Neural Recommendation Inference}. In
  \bibinfo{booktitle}{\emph{Proceedings of the ACM/IEEE Annual International
  Symposium on Computer Architecture}}.
\newblock


\bibitem[\protect\citeauthoryear{Gupta, Wu, Wang, Naumov, Reagen, Brooks,
  Cottel, Hazelwood, Hempstead, Jia, Lee, Malevich, Mudigere, Smelyanskiy,
  Xiong, and Zhang}{Gupta et~al\mbox{.}}{2020b}]%
        {gupta:2020:archimp}
\bibfield{author}{\bibinfo{person}{Udit Gupta}, \bibinfo{person}{Carole-Jean
  Wu}, \bibinfo{person}{Xiaodong Wang}, \bibinfo{person}{Maxim Naumov},
  \bibinfo{person}{Brandon Reagen}, \bibinfo{person}{David Brooks},
  \bibinfo{person}{Bradford Cottel}, \bibinfo{person}{Kim Hazelwood},
  \bibinfo{person}{Mark Hempstead}, \bibinfo{person}{Bill Jia},
  \bibinfo{person}{Hsien-Hsin~S. Lee}, \bibinfo{person}{Andrey Malevich},
  \bibinfo{person}{Dheevatsa Mudigere}, \bibinfo{person}{Mikhail Smelyanskiy},
  \bibinfo{person}{Liang Xiong}, {and} \bibinfo{person}{Xuan Zhang}.}
  \bibinfo{year}{2020}\natexlab{b}.
\newblock \showarticletitle{The Architectural Implications of Facebook's
  DNN-Based Personalized Recommendation}. In \bibinfo{booktitle}{\emph{2020
  IEEE International Symposium on High Performance Computer Architecture
  (HPCA)}}.
\newblock


\bibitem[\protect\citeauthoryear{{Gurobi Optimization, LLC}}{{Gurobi
  Optimization, LLC}}{2021}]%
        {gurobi}
\bibfield{author}{\bibinfo{person}{{Gurobi Optimization, LLC}}.}
  \bibinfo{year}{2021}\natexlab{}.
\newblock \showarticletitle{{Gurobi Optimizer Reference Manual}}.
\newblock  (\bibinfo{year}{2021}).
\newblock
\urldef\tempurl%
\url{https://www.gurobi.com}
\showURL{%
\tempurl}


\bibitem[\protect\citeauthoryear{Harris}{Harris}{2013}]%
        {nvidia:2013:uvm}
\bibfield{author}{\bibinfo{person}{Mark Harris}.}
  \bibinfo{year}{2013}\natexlab{}.
\newblock \bibinfo{title}{Unified Memory in {CUDA} 6}.
\newblock
\newblock
\newblock
\shownote{\url{https://developer.nvidia.com/blog/unified-memory-in-cuda-6/}}.


\bibitem[\protect\citeauthoryear{Hazelwood, Bird, Brooks, Chintala, Diril,
  Dzhulgakov, Fawzy, Jia, Jia, Kalro, Law, Lee, Lu, Noordhuis, Smelyanskiy,
  Xiong, and Wang}{Hazelwood et~al\mbox{.}}{2018}]%
        {hazelwood:2018:mlatfb}
\bibfield{author}{\bibinfo{person}{Kim Hazelwood}, \bibinfo{person}{Sarah
  Bird}, \bibinfo{person}{David Brooks}, \bibinfo{person}{Soumith Chintala},
  \bibinfo{person}{Utku Diril}, \bibinfo{person}{Dmytro Dzhulgakov},
  \bibinfo{person}{Mohamed Fawzy}, \bibinfo{person}{Bill Jia},
  \bibinfo{person}{Yangqing Jia}, \bibinfo{person}{Aditya Kalro},
  \bibinfo{person}{James Law}, \bibinfo{person}{Kevin Lee},
  \bibinfo{person}{Jason Lu}, \bibinfo{person}{Pieter Noordhuis},
  \bibinfo{person}{Misha Smelyanskiy}, \bibinfo{person}{Liang Xiong}, {and}
  \bibinfo{person}{Xiaodong Wang}.} \bibinfo{year}{2018}\natexlab{}.
\newblock \showarticletitle{Applied Machine Learning at Facebook: A Datacenter
  Infrastructure Perspective}. In \bibinfo{booktitle}{\emph{2018 IEEE
  International Symposium on High Performance Computer Architecture (HPCA)}}.
\newblock


\bibitem[\protect\citeauthoryear{He, Zhang, Ren, and Sun}{He
  et~al\mbox{.}}{2016}]%
        {resnet}
\bibfield{author}{\bibinfo{person}{Kaiming He}, \bibinfo{person}{Xiangyu
  Zhang}, \bibinfo{person}{Shaoqing Ren}, {and} \bibinfo{person}{Jian Sun}.}
  \bibinfo{year}{2016}\natexlab{}.
\newblock \showarticletitle{Deep Residual Learning for Image Recognition}. In
  \bibinfo{booktitle}{\emph{2016 IEEE Conference on Computer Vision and Pattern
  Recognition (CVPR)}}. \bibinfo{pages}{770--778}.
\newblock
\urldef\tempurl%
\url{https://doi.org/10.1109/CVPR.2016.90}
\showDOI{\tempurl}


\bibitem[\protect\citeauthoryear{Hsia, Gupta, Wilkening, Wu, Wei, and
  Brooks}{Hsia et~al\mbox{.}}{2020}]%
        {hsia:iiswc:2020}
\bibfield{author}{\bibinfo{person}{S. Hsia}, \bibinfo{person}{U. Gupta},
  \bibinfo{person}{M. Wilkening}, \bibinfo{person}{C. Wu}, \bibinfo{person}{G.
  Wei}, {and} \bibinfo{person}{D. Brooks}.} \bibinfo{year}{2020}\natexlab{}.
\newblock \showarticletitle{Cross-Stack Workload Characterization of Deep
  Recommendation Systems}. In \bibinfo{booktitle}{\emph{IEEE International
  Symposium on Workload Characterization (IISWC)}}. \bibinfo{publisher}{IEEE
  Computer Society}.
\newblock


\bibitem[\protect\citeauthoryear{Jiang, Deng, Yi, Hu, Zhou, Zheng, Huang, Guo,
  Wang, Song, Zhao, Wang, Sun, Zhang, Zhang, Li, Xu, Zhu, and Gai}{Jiang
  et~al\mbox{.}}{2019}]%
        {xdl}
\bibfield{author}{\bibinfo{person}{Biye Jiang}, \bibinfo{person}{Chao Deng},
  \bibinfo{person}{Huimin Yi}, \bibinfo{person}{Zelin Hu},
  \bibinfo{person}{Guorui Zhou}, \bibinfo{person}{Yang Zheng},
  \bibinfo{person}{Sui Huang}, \bibinfo{person}{Xinyang Guo},
  \bibinfo{person}{Dongyue Wang}, \bibinfo{person}{Yue Song},
  \bibinfo{person}{Liqin Zhao}, \bibinfo{person}{Zhi Wang},
  \bibinfo{person}{Peng Sun}, \bibinfo{person}{Yu Zhang}, \bibinfo{person}{Di
  Zhang}, \bibinfo{person}{Jinhui Li}, \bibinfo{person}{Jian Xu},
  \bibinfo{person}{Xiaoqiang Zhu}, {and} \bibinfo{person}{Kun Gai}.}
  \bibinfo{year}{2019}\natexlab{}.
\newblock \showarticletitle{XDL: An Industrial Deep Learning Framework for
  High-Dimensional Sparse Data}. In \bibinfo{booktitle}{\emph{Proceedings of
  the 1st International Workshop on Deep Learning Practice for High-Dimensional
  Sparse Data}} (Anchorage, Alaska) \emph{(\bibinfo{series}{DLP-KDD '19})}.
  \bibinfo{publisher}{Association for Computing Machinery},
  \bibinfo{address}{New York, NY, USA}, Article \bibinfo{articleno}{6},
  \bibinfo{numpages}{9}~pages.
\newblock
\showISBNx{9781450367837}
\urldef\tempurl%
\url{https://doi.org/10.1145/3326937.3341255}
\showDOI{\tempurl}


\bibitem[\protect\citeauthoryear{Jiang, Zhu, Lan, Yi, Cui, and Guo}{Jiang
  et~al\mbox{.}}{2020}]%
        {unified_arch_osdi}
\bibfield{author}{\bibinfo{person}{Yimin Jiang}, \bibinfo{person}{Yibo Zhu},
  \bibinfo{person}{Chang Lan}, \bibinfo{person}{Bairen Yi},
  \bibinfo{person}{Yong Cui}, {and} \bibinfo{person}{Chuanxiong Guo}.}
  \bibinfo{year}{2020}\natexlab{}.
\newblock \showarticletitle{A Unified Architecture for Accelerating Distributed
  {DNN} Training in Heterogeneous GPU/CPU Clusters}. In
  \bibinfo{booktitle}{\emph{14th {USENIX} Symposium on Operating Systems Design
  and Implementation ({OSDI} 20)}}. \bibinfo{publisher}{{USENIX} Association},
  \bibinfo{pages}{463--479}.
\newblock
\showISBNx{978-1-939133-19-9}
\urldef\tempurl%
\url{https://www.usenix.org/conference/osdi20/presentation/jiang}
\showURL{%
\tempurl}


\bibitem[\protect\citeauthoryear{Joglekar, Li, Chen, Xu, Wang, Adams, Khaitan,
  Liu, and Le}{Joglekar et~al\mbox{.}}{2020}]%
        {Joglekar:2020}
\bibfield{author}{\bibinfo{person}{Manas~R. Joglekar}, \bibinfo{person}{Cong
  Li}, \bibinfo{person}{Mei Chen}, \bibinfo{person}{Taibai Xu},
  \bibinfo{person}{Xiaoming Wang}, \bibinfo{person}{Jay~K. Adams},
  \bibinfo{person}{Pranav Khaitan}, \bibinfo{person}{Jiahui Liu}, {and}
  \bibinfo{person}{Quoc~V. Le}.} \bibinfo{year}{2020}\natexlab{}.
\newblock \showarticletitle{Neural Input Search for Large Scale Recommendation
  Models}. In \bibinfo{booktitle}{\emph{Proceedings of the 26th ACM SIGKDD
  International Conference on Knowledge Discovery \& Data Mining}} (Virtual
  Event, CA, USA) \emph{(\bibinfo{series}{KDD '20})}.
  \bibinfo{publisher}{Association for Computing Machinery},
  \bibinfo{address}{New York, NY, USA}, \bibinfo{pages}{2387–2397}.
\newblock
\showISBNx{9781450379984}
\urldef\tempurl%
\url{https://doi.org/10.1145/3394486.3403288}
\showDOI{\tempurl}


\bibitem[\protect\citeauthoryear{Jouppi, Young, Patil, Patterson, Agrawal,
  Bajwa, Bates, Bhatia, Boden, Borchers, Boyle, Cantin, Chao, Clark, Coriell,
  Daley, Dau, Dean, Gelb, Ghaemmaghami, Gottipati, Gulland, Hagmann, Ho,
  Hogberg, Hu, Hundt, Hurt, Ibarz, Jaffey, Jaworski, Kaplan, Khaitan,
  Killebrew, Koch, Kumar, Lacy, Laudon, Law, Le, Leary, Liu, Lucke, Lundin,
  MacKean, Maggiore, Mahony, Miller, Nagarajan, Narayanaswami, Ni, Nix, Norrie,
  Omernick, Penukonda, Phelps, Ross, Ross, Salek, Samadiani, Severn, Sizikov,
  Snelham, Souter, Steinberg, Swing, Tan, Thorson, Tian, Toma, Tuttle,
  Vasudevan, Walter, Wang, Wilcox, and Yoon}{Jouppi et~al\mbox{.}}{2017}]%
        {tpu}
\bibfield{author}{\bibinfo{person}{Norman~P. Jouppi}, \bibinfo{person}{Cliff
  Young}, \bibinfo{person}{Nishant Patil}, \bibinfo{person}{David Patterson},
  \bibinfo{person}{Gaurav Agrawal}, \bibinfo{person}{Raminder Bajwa},
  \bibinfo{person}{Sarah Bates}, \bibinfo{person}{Suresh Bhatia},
  \bibinfo{person}{Nan Boden}, \bibinfo{person}{Al Borchers},
  \bibinfo{person}{Rick Boyle}, \bibinfo{person}{Pierre-luc Cantin},
  \bibinfo{person}{Clifford Chao}, \bibinfo{person}{Chris Clark},
  \bibinfo{person}{Jeremy Coriell}, \bibinfo{person}{Mike Daley},
  \bibinfo{person}{Matt Dau}, \bibinfo{person}{Jeffrey Dean},
  \bibinfo{person}{Ben Gelb}, \bibinfo{person}{Tara~Vazir Ghaemmaghami},
  \bibinfo{person}{Rajendra Gottipati}, \bibinfo{person}{William Gulland},
  \bibinfo{person}{Robert Hagmann}, \bibinfo{person}{C.~Richard Ho},
  \bibinfo{person}{Doug Hogberg}, \bibinfo{person}{John Hu},
  \bibinfo{person}{Robert Hundt}, \bibinfo{person}{Dan Hurt},
  \bibinfo{person}{Julian Ibarz}, \bibinfo{person}{Aaron Jaffey},
  \bibinfo{person}{Alek Jaworski}, \bibinfo{person}{Alexander Kaplan},
  \bibinfo{person}{Harshit Khaitan}, \bibinfo{person}{Daniel Killebrew},
  \bibinfo{person}{Andy Koch}, \bibinfo{person}{Naveen Kumar},
  \bibinfo{person}{Steve Lacy}, \bibinfo{person}{James Laudon},
  \bibinfo{person}{James Law}, \bibinfo{person}{Diemthu Le},
  \bibinfo{person}{Chris Leary}, \bibinfo{person}{Zhuyuan Liu},
  \bibinfo{person}{Kyle Lucke}, \bibinfo{person}{Alan Lundin},
  \bibinfo{person}{Gordon MacKean}, \bibinfo{person}{Adriana Maggiore},
  \bibinfo{person}{Maire Mahony}, \bibinfo{person}{Kieran Miller},
  \bibinfo{person}{Rahul Nagarajan}, \bibinfo{person}{Ravi Narayanaswami},
  \bibinfo{person}{Ray Ni}, \bibinfo{person}{Kathy Nix},
  \bibinfo{person}{Thomas Norrie}, \bibinfo{person}{Mark Omernick},
  \bibinfo{person}{Narayana Penukonda}, \bibinfo{person}{Andy Phelps},
  \bibinfo{person}{Jonathan Ross}, \bibinfo{person}{Matt Ross},
  \bibinfo{person}{Amir Salek}, \bibinfo{person}{Emad Samadiani},
  \bibinfo{person}{Chris Severn}, \bibinfo{person}{Gregory Sizikov},
  \bibinfo{person}{Matthew Snelham}, \bibinfo{person}{Jed Souter},
  \bibinfo{person}{Dan Steinberg}, \bibinfo{person}{Andy Swing},
  \bibinfo{person}{Mercedes Tan}, \bibinfo{person}{Gregory Thorson},
  \bibinfo{person}{Bo Tian}, \bibinfo{person}{Horia Toma},
  \bibinfo{person}{Erick Tuttle}, \bibinfo{person}{Vijay Vasudevan},
  \bibinfo{person}{Richard Walter}, \bibinfo{person}{Walter Wang},
  \bibinfo{person}{Eric Wilcox}, {and} \bibinfo{person}{Doe~Hyun Yoon}.}
  \bibinfo{year}{2017}\natexlab{}.
\newblock \showarticletitle{In-datacenter performance analysis of a tensor
  processing unit}. In \bibinfo{booktitle}{\emph{Proceedings of the ACM/IEEE
  44th Annual International Symposium on Computer Architecture}}.
\newblock


\bibitem[\protect\citeauthoryear{Jumper, Evans, Pritzel, Green, Figurnov,
  Ronneberger, Tunyasuvunakool, Bates, Žídek, Potapenko, Bridgland, Meyer,
  Kohl, Ballard, Cowie, Romera-Paredes, Nikolov, Jain, Adler, Back, Petersen,
  Reiman, Clancy, Zielinski, Steinegger, Pacholska, Berghammer, Bodenstein,
  Silver, Vinyals, Senior, Kavukcuoglu, Kohli, and Hassabis}{Jumper
  et~al\mbox{.}}{2021}]%
        {AlphaFold}
\bibfield{author}{\bibinfo{person}{John Jumper}, \bibinfo{person}{Richard
  Evans}, \bibinfo{person}{Alexander Pritzel}, \bibinfo{person}{Tim Green},
  \bibinfo{person}{Michael Figurnov}, \bibinfo{person}{Olaf Ronneberger},
  \bibinfo{person}{Kathryn Tunyasuvunakool}, \bibinfo{person}{Russ Bates},
  \bibinfo{person}{Augustin Žídek}, \bibinfo{person}{Anna Potapenko},
  \bibinfo{person}{Alex Bridgland}, \bibinfo{person}{Clemens Meyer},
  \bibinfo{person}{Simon A.~A. Kohl}, \bibinfo{person}{Andrew~J. Ballard},
  \bibinfo{person}{Andrew Cowie}, \bibinfo{person}{Bernardino Romera-Paredes},
  \bibinfo{person}{Stanislav Nikolov}, \bibinfo{person}{Rishub Jain},
  \bibinfo{person}{Jonas Adler}, \bibinfo{person}{Trevor Back},
  \bibinfo{person}{Stig Petersen}, \bibinfo{person}{David Reiman},
  \bibinfo{person}{Ellen Clancy}, \bibinfo{person}{Michal Zielinski},
  \bibinfo{person}{Martin Steinegger}, \bibinfo{person}{Michalina Pacholska},
  \bibinfo{person}{Tamas Berghammer}, \bibinfo{person}{Sebastian Bodenstein},
  \bibinfo{person}{David Silver}, \bibinfo{person}{Oriol Vinyals},
  \bibinfo{person}{Andrew~W. Senior}, \bibinfo{person}{Koray Kavukcuoglu},
  \bibinfo{person}{Pushmeet Kohli}, {and} \bibinfo{person}{Demis Hassabis}.}
  \bibinfo{year}{2021}\natexlab{}.
\newblock \showarticletitle{Highly accurate protein structure prediction with
  AlphaFold}.
\newblock \bibinfo{journal}{\emph{Nature}} (\bibinfo{year}{2021}).
\newblock


\bibitem[\protect\citeauthoryear{Kang, Cheng, Yao, Yi, Chen, Hong, and
  Chi}{Kang et~al\mbox{.}}{2021}]%
        {kang2021learning}
\bibfield{author}{\bibinfo{person}{Wang-Cheng Kang},
  \bibinfo{person}{Derek~Zhiyuan Cheng}, \bibinfo{person}{Tiansheng Yao},
  \bibinfo{person}{Xinyang Yi}, \bibinfo{person}{Ting Chen},
  \bibinfo{person}{Lichan Hong}, {and} \bibinfo{person}{Ed~H. Chi}.}
  \bibinfo{year}{2021}\natexlab{}.
\newblock \showarticletitle{Learning to Embed Categorical Features without
  Embedding Tables for Recommendation}.
\newblock \bibinfo{journal}{\emph{CoRR}} (\bibinfo{year}{2021}).
\newblock
\urldef\tempurl%
\url{https://arxiv.org/abs/2010.10784}
\showURL{%
\tempurl}


\bibitem[\protect\citeauthoryear{Ke, Gupta, Cho, Brooks, Chandra, Diril,
  Firoozshahian, Hazelwood, Jia, Lee, Li, Maher, Mudigere, Naumov, Schatz,
  Smelyanskiy, Wang, Reagen, Wu, Hempstead, and Zhang}{Ke
  et~al\mbox{.}}{2020}]%
        {rec-nmp}
\bibfield{author}{\bibinfo{person}{Liu Ke}, \bibinfo{person}{Udit Gupta},
  \bibinfo{person}{Benjamin~Youngjae Cho}, \bibinfo{person}{David Brooks},
  \bibinfo{person}{Vikas Chandra}, \bibinfo{person}{Utku Diril},
  \bibinfo{person}{Amin Firoozshahian}, \bibinfo{person}{Kim~M. Hazelwood},
  \bibinfo{person}{Bill Jia}, \bibinfo{person}{Hsien{-}Hsin~S. Lee},
  \bibinfo{person}{Meng Li}, \bibinfo{person}{Bert Maher},
  \bibinfo{person}{Dheevatsa Mudigere}, \bibinfo{person}{Maxim Naumov},
  \bibinfo{person}{Martin Schatz}, \bibinfo{person}{Mikhail Smelyanskiy},
  \bibinfo{person}{Xiaodong Wang}, \bibinfo{person}{Brandon Reagen},
  \bibinfo{person}{Carole{-}Jean Wu}, \bibinfo{person}{Mark Hempstead}, {and}
  \bibinfo{person}{Xuan Zhang}.} \bibinfo{year}{2020}\natexlab{}.
\newblock \showarticletitle{RecNMP: Accelerating Personalized Recommendation
  with Near-Memory Processing}. In \bibinfo{booktitle}{\emph{47th {ACM/IEEE}
  Annual International Symposium on Computer Architecture, {ISCA} 2020,
  Valencia, Spain, May 30 - June 3, 2020}}. \bibinfo{publisher}{{IEEE}},
  \bibinfo{pages}{790--803}.
\newblock
\urldef\tempurl%
\url{https://doi.org/10.1109/ISCA45697.2020.00070}
\showDOI{\tempurl}


\bibitem[\protect\citeauthoryear{Kumar, Bradbury, Young, Wang, Levskaya,
  Hechtman, Chen, Lee, Deveci, Kumar, Kanwar, Wang, Wanderman-Milne, Lacy,
  Wang, Oguntebi, Zu, Xu, and Swing}{Kumar et~al\mbox{.}}{2021}]%
        {kumar2021exploring}
\bibfield{author}{\bibinfo{person}{Sameer Kumar}, \bibinfo{person}{James
  Bradbury}, \bibinfo{person}{Cliff Young}, \bibinfo{person}{Yu~Emma Wang},
  \bibinfo{person}{Anselm Levskaya}, \bibinfo{person}{Blake Hechtman},
  \bibinfo{person}{Dehao Chen}, \bibinfo{person}{HyoukJoong Lee},
  \bibinfo{person}{Mehmet Deveci}, \bibinfo{person}{Naveen Kumar},
  \bibinfo{person}{Pankaj Kanwar}, \bibinfo{person}{Shibo Wang},
  \bibinfo{person}{Skye Wanderman-Milne}, \bibinfo{person}{Steve Lacy},
  \bibinfo{person}{Tao Wang}, \bibinfo{person}{Tayo Oguntebi},
  \bibinfo{person}{Yazhou Zu}, \bibinfo{person}{Yuanzhong Xu}, {and}
  \bibinfo{person}{Andy Swing}.} \bibinfo{year}{2021}\natexlab{}.
\newblock \showarticletitle{Exploring the limits of Concurrency in ML Training
  on Google TPUs}.
\newblock
\showeprint[arxiv]{2011.03641}~[cs.LG]


\bibitem[\protect\citeauthoryear{Liu, Zhao, Wang, Liu, and Tang}{Liu
  et~al\mbox{.}}{2020}]%
        {Liu:2020}
\bibfield{author}{\bibinfo{person}{Haochen Liu}, \bibinfo{person}{Xiangyu
  Zhao}, \bibinfo{person}{Chong Wang}, \bibinfo{person}{Xiaobing Liu}, {and}
  \bibinfo{person}{Jiliang Tang}.} \bibinfo{year}{2020}\natexlab{}.
\newblock \showarticletitle{Automated Embedding Size Search in Deep Recommender
  Systems}. In \bibinfo{booktitle}{\emph{Proceedings of the 43rd International
  ACM SIGIR Conference on Research and Development in Information Retrieval}}
  (Virtual Event, China) \emph{(\bibinfo{series}{SIGIR '20})}.
  \bibinfo{publisher}{Association for Computing Machinery},
  \bibinfo{address}{New York, NY, USA}, \bibinfo{pages}{2307–2316}.
\newblock
\showISBNx{9781450380164}
\urldef\tempurl%
\url{https://doi.org/10.1145/3397271.3401436}
\showDOI{\tempurl}


\bibitem[\protect\citeauthoryear{Lui, Yetim, Özkan, Zhao, Tsai, Wu, and
  Hempstead}{Lui et~al\mbox{.}}{2021}]%
        {lui:2021:capacity}
\bibfield{author}{\bibinfo{person}{Michael Lui}, \bibinfo{person}{Yavuz Yetim},
  \bibinfo{person}{Özgür Özkan}, \bibinfo{person}{Zhuoran Zhao},
  \bibinfo{person}{Shin-Yeh Tsai}, \bibinfo{person}{Carole-Jean Wu}, {and}
  \bibinfo{person}{Mark Hempstead}.} \bibinfo{year}{2021}\natexlab{}.
\newblock \showarticletitle{Understanding Capacity-Driven Scale-Out Neural
  Recommendation Inference}. In \bibinfo{booktitle}{\emph{2021 IEEE
  International Symposium on Performance Analysis of Systems and Software
  (ISPASS)}}.
\newblock


\bibitem[\protect\citeauthoryear{Lutz, Bre\ss{}, Zeuch, Rabl, and Markl}{Lutz
  et~al\mbox{.}}{2020}]%
        {pump_up_sigmod}
\bibfield{author}{\bibinfo{person}{Clemens Lutz}, \bibinfo{person}{Sebastian
  Bre\ss{}}, \bibinfo{person}{Steffen Zeuch}, \bibinfo{person}{Tilmann Rabl},
  {and} \bibinfo{person}{Volker Markl}.} \bibinfo{year}{2020}\natexlab{}.
\newblock \showarticletitle{Pump Up the Volume: Processing Large Data on GPUs
  with Fast Interconnects}. In \bibinfo{booktitle}{\emph{Proceedings of the
  2020 ACM SIGMOD International Conference on Management of Data}} (Portland,
  OR, USA) \emph{(\bibinfo{series}{SIGMOD '20})}.
  \bibinfo{publisher}{Association for Computing Machinery},
  \bibinfo{address}{New York, NY, USA}, \bibinfo{pages}{1633–1649}.
\newblock
\showISBNx{9781450367356}
\urldef\tempurl%
\url{https://doi.org/10.1145/3318464.3389705}
\showDOI{\tempurl}


\bibitem[\protect\citeauthoryear{Maeng, Bharuka, Gao, Jeffrey, Saraph, Su,
  Trippel, Yang, Rabbat, Lucia, and Wu}{Maeng et~al\mbox{.}}{2021}]%
        {maeng:2021:cpr}
\bibfield{author}{\bibinfo{person}{Kiwan Maeng}, \bibinfo{person}{Shivam
  Bharuka}, \bibinfo{person}{Isabel Gao}, \bibinfo{person}{Mark Jeffrey},
  \bibinfo{person}{Vikram Saraph}, \bibinfo{person}{Bor-Yiing Su},
  \bibinfo{person}{Caroline Trippel}, \bibinfo{person}{Jiyan Yang},
  \bibinfo{person}{Mike Rabbat}, \bibinfo{person}{Brandon Lucia}, {and}
  \bibinfo{person}{Carole-Jean Wu}.} \bibinfo{year}{2021}\natexlab{}.
\newblock \showarticletitle{Understanding and Improving Failure Tolerant
  Training for Deep Learning Recommendation with Partial Recovery}. In
  \bibinfo{booktitle}{\emph{Proceedings of Machine Learning and Systems}}.
\newblock


\bibitem[\protect\citeauthoryear{Mattson, Cheng, Diamos, Coleman, Micikevicius,
  Patterson, Tang, Wei, Bailis, Bittorf, Brooks, Chen, Dutta, Gupta, Hazelwood,
  Hock, Huang, Kang, Kanter, Kumar, Liao, Narayanan, Oguntebi, Pekhimenko,
  Pentecost, Janapa~Reddi, Robie, St~John, Wu, Xu, Young, and Zaharia}{Mattson
  et~al\mbox{.}}{2020}]%
        {mattson:2020:mlperftraining}
\bibfield{author}{\bibinfo{person}{Peter Mattson}, \bibinfo{person}{Christine
  Cheng}, \bibinfo{person}{Gregory Diamos}, \bibinfo{person}{Cody Coleman},
  \bibinfo{person}{Paulius Micikevicius}, \bibinfo{person}{David Patterson},
  \bibinfo{person}{Hanlin Tang}, \bibinfo{person}{Gu-Yeon Wei},
  \bibinfo{person}{Peter Bailis}, \bibinfo{person}{Victor Bittorf},
  \bibinfo{person}{David Brooks}, \bibinfo{person}{Dehao Chen},
  \bibinfo{person}{Debo Dutta}, \bibinfo{person}{Udit Gupta},
  \bibinfo{person}{Kim Hazelwood}, \bibinfo{person}{Andy Hock},
  \bibinfo{person}{Xinyuan Huang}, \bibinfo{person}{Daniel Kang},
  \bibinfo{person}{David Kanter}, \bibinfo{person}{Naveen Kumar},
  \bibinfo{person}{Jeffery Liao}, \bibinfo{person}{Deepak Narayanan},
  \bibinfo{person}{Tayo Oguntebi}, \bibinfo{person}{Gennady Pekhimenko},
  \bibinfo{person}{Lillian Pentecost}, \bibinfo{person}{Vijay Janapa~Reddi},
  \bibinfo{person}{Taylor Robie}, \bibinfo{person}{Tom St~John},
  \bibinfo{person}{Carole-Jean Wu}, \bibinfo{person}{Lingjie Xu},
  \bibinfo{person}{Cliff Young}, {and} \bibinfo{person}{Matei Zaharia}.}
  \bibinfo{year}{2020}\natexlab{}.
\newblock \showarticletitle{MLPerf Training Benchmark}. In
  \bibinfo{booktitle}{\emph{Proceedings of Machine Learning and Systems}}.
\newblock


\bibitem[\protect\citeauthoryear{Min, Mailthody, Qureshi, Xiong, Ebrahimi, and
  Hwu}{Min et~al\mbox{.}}{2020}]%
        {emogi}
\bibfield{author}{\bibinfo{person}{Seung~Won Min},
  \bibinfo{person}{Vikram~Sharma Mailthody}, \bibinfo{person}{Zaid Qureshi},
  \bibinfo{person}{Jinjun Xiong}, \bibinfo{person}{Eiman Ebrahimi}, {and}
  \bibinfo{person}{Wen-mei Hwu}.} \bibinfo{year}{2020}\natexlab{}.
\newblock \showarticletitle{EMOGI: Efficient Memory-Access for out-of-Memory
  Graph-Traversal in GPUs}.
\newblock \bibinfo{journal}{\emph{Proc. VLDB Endow.}} \bibinfo{volume}{14},
  \bibinfo{number}{2} (\bibinfo{date}{Oct.} \bibinfo{year}{2020}),
  \bibinfo{pages}{114–127}.
\newblock
\showISSN{2150-8097}
\urldef\tempurl%
\url{https://doi.org/10.14778/3425879.3425883}
\showDOI{\tempurl}


\bibitem[\protect\citeauthoryear{Mudigere, Hao, Huang, Tulloch, Sridharan, Liu,
  Ozdal, Nie, Park, Luo, Yang, Gao, Ivchenko, Basant, Hu, Yang, Ardestani,
  Wang, Komuravelli, Chu, Yilmaz, Li, Qian, Feng, Ma, Yang, Wen, Li, Yang, Sun,
  Zhao, Melts, Dhulipala, Kishore, Graf, Eisenman, Matam, Gangidi, Chen,
  Krishnan, Nayak, Nair, Muthiah, khorashadi, Bhattacharya, Lapukhov, Naumov,
  Qiao, Smelyanskiy, Jia, and Rao}{Mudigere et~al\mbox{.}}{2021}]%
        {mudigere2021highperformance}
\bibfield{author}{\bibinfo{person}{Dheevatsa Mudigere}, \bibinfo{person}{Yuchen
  Hao}, \bibinfo{person}{Jianyu Huang}, \bibinfo{person}{Andrew Tulloch},
  \bibinfo{person}{Srinivas Sridharan}, \bibinfo{person}{Xing Liu},
  \bibinfo{person}{Mustafa Ozdal}, \bibinfo{person}{Jade Nie},
  \bibinfo{person}{Jongsoo Park}, \bibinfo{person}{Liang Luo},
  \bibinfo{person}{Jie~Amy Yang}, \bibinfo{person}{Leon Gao},
  \bibinfo{person}{Dmytro Ivchenko}, \bibinfo{person}{Aarti Basant},
  \bibinfo{person}{Yuxi Hu}, \bibinfo{person}{Jiyan Yang},
  \bibinfo{person}{Ehsan~K. Ardestani}, \bibinfo{person}{Xiaodong Wang},
  \bibinfo{person}{Rakesh Komuravelli}, \bibinfo{person}{Ching-Hsiang Chu},
  \bibinfo{person}{Serhat Yilmaz}, \bibinfo{person}{Huayu Li},
  \bibinfo{person}{Jiyuan Qian}, \bibinfo{person}{Zhuobo Feng},
  \bibinfo{person}{Yinbin Ma}, \bibinfo{person}{Junjie Yang},
  \bibinfo{person}{Ellie Wen}, \bibinfo{person}{Hong Li}, \bibinfo{person}{Lin
  Yang}, \bibinfo{person}{Chonglin Sun}, \bibinfo{person}{Whitney Zhao},
  \bibinfo{person}{Dimitry Melts}, \bibinfo{person}{Krishna Dhulipala},
  \bibinfo{person}{KR Kishore}, \bibinfo{person}{Tyler Graf},
  \bibinfo{person}{Assaf Eisenman}, \bibinfo{person}{Kiran~Kumar Matam},
  \bibinfo{person}{Adi Gangidi}, \bibinfo{person}{Guoqiang~Jerry Chen},
  \bibinfo{person}{Manoj Krishnan}, \bibinfo{person}{Avinash Nayak},
  \bibinfo{person}{Krishnakumar Nair}, \bibinfo{person}{Bharath Muthiah},
  \bibinfo{person}{Mahmoud khorashadi}, \bibinfo{person}{Pallab Bhattacharya},
  \bibinfo{person}{Petr Lapukhov}, \bibinfo{person}{Maxim Naumov},
  \bibinfo{person}{Lin Qiao}, \bibinfo{person}{Mikhail Smelyanskiy},
  \bibinfo{person}{Bill Jia}, {and} \bibinfo{person}{Vijay Rao}.}
  \bibinfo{year}{2021}\natexlab{}.
\newblock \showarticletitle{High-performance, Distributed Training of
  Large-scale Deep Learning Recommendation Models}.
\newblock \bibinfo{journal}{\emph{CoRR}} (\bibinfo{year}{2021}).
\newblock
\showeprint[arxiv]{2104.05158}~[cs.DC]


\bibitem[\protect\citeauthoryear{Naumov, Kim, Mudigere, Sridharan, Wang, Zhao,
  Yilmaz, Kim, Yuen, Ozdal, Nair, Gao, Su, Yang, and Smelyanskiy}{Naumov
  et~al\mbox{.}}{2020}]%
        {naumov2020deep}
\bibfield{author}{\bibinfo{person}{Maxim Naumov}, \bibinfo{person}{John Kim},
  \bibinfo{person}{Dheevatsa Mudigere}, \bibinfo{person}{Srinivas Sridharan},
  \bibinfo{person}{Xiaodong Wang}, \bibinfo{person}{Whitney Zhao},
  \bibinfo{person}{Serhat Yilmaz}, \bibinfo{person}{Changkyu Kim},
  \bibinfo{person}{Hector Yuen}, \bibinfo{person}{Mustafa Ozdal},
  \bibinfo{person}{Krishnakumar Nair}, \bibinfo{person}{Isabel Gao},
  \bibinfo{person}{Bor-Yiing Su}, \bibinfo{person}{Jiyan Yang}, {and}
  \bibinfo{person}{Mikhail Smelyanskiy}.} \bibinfo{year}{2020}\natexlab{}.
\newblock \showarticletitle{Deep Learning Training in Facebook Data Centers:
  Design of Scale-up and Scale-out Systems}.
\newblock \bibinfo{journal}{\emph{CoRR}} (\bibinfo{year}{2020}).
\newblock
\showeprint[arxiv]{2003.09518}~[cs.DC]


\bibitem[\protect\citeauthoryear{Naumov, Mudigere, Shi, Huang, Sundaraman,
  Park, Wang, Gupta, Wu, Azzolini, Dzhulgakov, Mallevich, Cherniavskii, Lu,
  Krishnamoorthi, Yu, Kondratenko, Pereira, Chen, Chen, Rao, Jia, Xiong, and
  Smelyanskiy}{Naumov et~al\mbox{.}}{2019}]%
        {naumov2019deep}
\bibfield{author}{\bibinfo{person}{Maxim Naumov}, \bibinfo{person}{Dheevatsa
  Mudigere}, \bibinfo{person}{Hao-Jun~Michael Shi}, \bibinfo{person}{Jianyu
  Huang}, \bibinfo{person}{Narayanan Sundaraman}, \bibinfo{person}{Jongsoo
  Park}, \bibinfo{person}{Xiaodong Wang}, \bibinfo{person}{Udit Gupta},
  \bibinfo{person}{Carole-Jean Wu}, \bibinfo{person}{Alisson~G. Azzolini},
  \bibinfo{person}{Dmytro Dzhulgakov}, \bibinfo{person}{Andrey Mallevich},
  \bibinfo{person}{Ilia Cherniavskii}, \bibinfo{person}{Yinghai Lu},
  \bibinfo{person}{Raghuraman Krishnamoorthi}, \bibinfo{person}{Ansha Yu},
  \bibinfo{person}{Volodymyr Kondratenko}, \bibinfo{person}{Stephanie Pereira},
  \bibinfo{person}{Xianjie Chen}, \bibinfo{person}{Wenlin Chen},
  \bibinfo{person}{Vijay Rao}, \bibinfo{person}{Bill Jia},
  \bibinfo{person}{Liang Xiong}, {and} \bibinfo{person}{Misha Smelyanskiy}.}
  \bibinfo{year}{2019}\natexlab{}.
\newblock \showarticletitle{Deep Learning Recommendation Model for
  Personalization and Recommendation Systems}.
\newblock  (\bibinfo{year}{2019}).
\newblock
\showeprint[arxiv]{1906.00091}~[cs.IR]


\bibitem[\protect\citeauthoryear{Raimond}{Raimond}{2018}]%
        {Raimond:netflix}
\bibfield{author}{\bibinfo{person}{Yves Raimond}.}
  \bibinfo{year}{2018}\natexlab{}.
\newblock \bibinfo{title}{Deep Learning for Recommender Systems}.
\newblock
\newblock
\newblock
\shownote{\url{https://www.slideshare.net/moustaki/deep-learning-for-
  recommender-systems-86752234}}.


\bibitem[\protect\citeauthoryear{Reddi, Cheng, Kanter, Mattson, Schmuelling,
  Wu, Anderson, Breughe, Charlebois, Chou, Chukka, Coleman, Davis, Deng,
  Diamos, Duke, Fick, Gardner, Hubara, Idgunji, Jablin, Jiao, John, Kanwar,
  Lee, Liao, Lokhmotov, Massa, Meng, Micikevicius, Osborne, Pekhimenko, Rajan,
  Sequeira, Sirasao, Sun, Tang, Thomson, Wei, Wu, Xu, Yamada, Yu, Yuan, Zhong,
  Zhang, and Zhou}{Reddi et~al\mbox{.}}{2020}]%
        {reddi:2020:mlpefinference}
\bibfield{author}{\bibinfo{person}{Vijay~Janapa Reddi},
  \bibinfo{person}{Christine Cheng}, \bibinfo{person}{David Kanter},
  \bibinfo{person}{Peter Mattson}, \bibinfo{person}{Guenther Schmuelling},
  \bibinfo{person}{Carole-Jean Wu}, \bibinfo{person}{Brian Anderson},
  \bibinfo{person}{Maximilien Breughe}, \bibinfo{person}{Mark Charlebois},
  \bibinfo{person}{William Chou}, \bibinfo{person}{Ramesh Chukka},
  \bibinfo{person}{Cody Coleman}, \bibinfo{person}{Sam Davis},
  \bibinfo{person}{Pan Deng}, \bibinfo{person}{Greg Diamos},
  \bibinfo{person}{Jared Duke}, \bibinfo{person}{Dave Fick},
  \bibinfo{person}{J.~Scott Gardner}, \bibinfo{person}{Itay Hubara},
  \bibinfo{person}{Sachin Idgunji}, \bibinfo{person}{Thomas~B. Jablin},
  \bibinfo{person}{Jeff Jiao}, \bibinfo{person}{Tom~St. John},
  \bibinfo{person}{Pankaj Kanwar}, \bibinfo{person}{David Lee},
  \bibinfo{person}{Jeffery Liao}, \bibinfo{person}{Anton Lokhmotov},
  \bibinfo{person}{Francisco Massa}, \bibinfo{person}{Peng Meng},
  \bibinfo{person}{Paulius Micikevicius}, \bibinfo{person}{Colin Osborne},
  \bibinfo{person}{Gennady Pekhimenko}, \bibinfo{person}{Arun Tejusve~Raghunath
  Rajan}, \bibinfo{person}{Dilip Sequeira}, \bibinfo{person}{Ashish Sirasao},
  \bibinfo{person}{Fei Sun}, \bibinfo{person}{Hanlin Tang},
  \bibinfo{person}{Michael Thomson}, \bibinfo{person}{Frank Wei},
  \bibinfo{person}{Ephrem Wu}, \bibinfo{person}{Lingjie Xu},
  \bibinfo{person}{Koichi Yamada}, \bibinfo{person}{Bing Yu},
  \bibinfo{person}{George Yuan}, \bibinfo{person}{Aaron Zhong},
  \bibinfo{person}{Peizhao Zhang}, {and} \bibinfo{person}{Yuchen Zhou}.}
  \bibinfo{year}{2020}\natexlab{}.
\newblock \showarticletitle{MLPerf Inference Benchmark}. In
  \bibinfo{booktitle}{\emph{2020 ACM/IEEE 47th Annual International Symposium
  on Computer Architecture (ISCA)}}.
\newblock


\bibitem[\protect\citeauthoryear{Sullivan}{Sullivan}{2016}]%
        {googe:2016:rankbrain}
\bibfield{author}{\bibinfo{person}{Danny Sullivan}.}
  \bibinfo{year}{2016}\natexlab{}.
\newblock \bibinfo{title}{Google uses {RankBrain} for every search, impacts
  rankings of ``lots'' of them}.
\newblock
\newblock
\newblock
\shownote{\url{https://searchengineland.com/google-loves-rankbrain-uses-for-every-search-252526}}.


\bibitem[\protect\citeauthoryear{Vaswani, Shazeer, Parmar, Uszkoreit, Jones,
  Gomez, Kaiser, and Polosukhin}{Vaswani et~al\mbox{.}}{2017}]%
        {transformers}
\bibfield{author}{\bibinfo{person}{Ashish Vaswani}, \bibinfo{person}{Noam
  Shazeer}, \bibinfo{person}{Niki Parmar}, \bibinfo{person}{Jakob Uszkoreit},
  \bibinfo{person}{Llion Jones}, \bibinfo{person}{Aidan~N. Gomez},
  \bibinfo{person}{undefinedukasz Kaiser}, {and} \bibinfo{person}{Illia
  Polosukhin}.} \bibinfo{year}{2017}\natexlab{}.
\newblock \showarticletitle{Attention is All You Need}. In
  \bibinfo{booktitle}{\emph{Proceedings of the 31st International Conference on
  Neural Information Processing Systems}} (Long Beach, California, USA)
  \emph{(\bibinfo{series}{NIPS'17})}. \bibinfo{publisher}{Curran Associates
  Inc.}, \bibinfo{address}{Red Hook, NY, USA}, \bibinfo{pages}{6000–6010}.
\newblock
\showISBNx{9781510860964}


\bibitem[\protect\citeauthoryear{Wang, Wu, Wang, Hazelwood, and Brooks}{Wang
  et~al\mbox{.}}{2021}]%
        {wang2020exploiting}
\bibfield{author}{\bibinfo{person}{Yu~Emma Wang}, \bibinfo{person}{Carole-Jean
  Wu}, \bibinfo{person}{Xiaodong Wang}, \bibinfo{person}{Kim Hazelwood}, {and}
  \bibinfo{person}{David Brooks}.} \bibinfo{year}{2021}\natexlab{}.
\newblock \showarticletitle{Exploiting Parallelism Opportunities with Deep
  Learning Frameworks}.
\newblock \bibinfo{journal}{\emph{ACM Transactions on Architecture and Code
  Optimization}} \bibinfo{volume}{18}, \bibinfo{number}{1}
  (\bibinfo{year}{2021}).
\newblock


\bibitem[\protect\citeauthoryear{Weinberger, Dasgupta, Langford, Smola, and
  Attenberg}{Weinberger et~al\mbox{.}}{2009}]%
        {hashing_trick}
\bibfield{author}{\bibinfo{person}{Kilian Weinberger}, \bibinfo{person}{Anirban
  Dasgupta}, \bibinfo{person}{John Langford}, \bibinfo{person}{Alex Smola},
  {and} \bibinfo{person}{Josh Attenberg}.} \bibinfo{year}{2009}\natexlab{}.
\newblock \showarticletitle{Feature Hashing for Large Scale Multitask
  Learning}. In \bibinfo{booktitle}{\emph{Proceedings of the 26th Annual
  International Conference on Machine Learning}} (Montreal, Quebec, Canada)
  \emph{(\bibinfo{series}{ICML '09})}. \bibinfo{publisher}{Association for
  Computing Machinery}, \bibinfo{address}{New York, NY, USA},
  \bibinfo{pages}{1113–1120}.
\newblock
\showISBNx{9781605585161}
\urldef\tempurl%
\url{https://doi.org/10.1145/1553374.1553516}
\showDOI{\tempurl}


\bibitem[\protect\citeauthoryear{Weyn, Durran, and Caruana}{Weyn
  et~al\mbox{.}}{2020}]%
        {Weyn_2020}
\bibfield{author}{\bibinfo{person}{Jonathan~A. Weyn}, \bibinfo{person}{Dale~R.
  Durran}, {and} \bibinfo{person}{Rich Caruana}.}
  \bibinfo{year}{2020}\natexlab{}.
\newblock \showarticletitle{Improving Data‐Driven Global Weather Prediction
  Using Deep Convolutional Neural Networks on a Cubed Sphere}.
\newblock \bibinfo{journal}{\emph{Journal of Advances in Modeling Earth
  Systems}} \bibinfo{volume}{12}, \bibinfo{number}{9} (\bibinfo{date}{Sep}
  \bibinfo{year}{2020}).
\newblock
\showISSN{1942-2466}
\urldef\tempurl%
\url{https://doi.org/10.1029/2020ms002109}
\showDOI{\tempurl}


\bibitem[\protect\citeauthoryear{Wilkening, Gupta, Hsia, Trippel, Wu, Brooks,
  and Wei}{Wilkening et~al\mbox{.}}{2021}]%
        {wilkening:2021:recssd}
\bibfield{author}{\bibinfo{person}{Mark Wilkening}, \bibinfo{person}{Udit
  Gupta}, \bibinfo{person}{Samuel Hsia}, \bibinfo{person}{Caroline Trippel},
  \bibinfo{person}{Carole-Jean Wu}, \bibinfo{person}{David Brooks}, {and}
  \bibinfo{person}{Gu-Yeon Wei}.} \bibinfo{year}{2021}\natexlab{}.
\newblock \showarticletitle{RecSSD: Near Data Processing for Solid State Drive
  Based Recommendation Inference}. In \bibinfo{booktitle}{\emph{Proceedings of
  the 26th ACM International Conference on Architectural Support for
  Programming Languages and Operating Systems}}.
\newblock


\bibitem[\protect\citeauthoryear{Wu, Burke, Chi, Konstan, McAuley, Raimond, and
  Zhang}{Wu et~al\mbox{.}}{2020}]%
        {wu:2020:mlperfrec}
\bibfield{author}{\bibinfo{person}{Carole{-}Jean Wu}, \bibinfo{person}{Robin
  Burke}, \bibinfo{person}{Ed Chi}, \bibinfo{person}{Joseph~A. Konstan},
  \bibinfo{person}{Julian~J. McAuley}, \bibinfo{person}{Yves Raimond}, {and}
  \bibinfo{person}{Hao Zhang}.} \bibinfo{year}{2020}\natexlab{}.
\newblock \showarticletitle{Developing a Recommendation Benchmark for MLPerf
  Training and Inference}.
\newblock \bibinfo{journal}{\emph{CoRR}}  \bibinfo{volume}{abs/2003.07336}
  (\bibinfo{year}{2020}).
\newblock
\urldef\tempurl%
\url{https://arxiv.org/abs/2003.07336}
\showURL{%
\tempurl}


\bibitem[\protect\citeauthoryear{Wu, Raghavendra, Gupta, Acun, Ardalani, Maeng,
  Chang, Behram, Huang, Bai, Gschwind, Gupta, Ott, Melnikov, Candido, Brooks,
  Chauhan, Lee, Lee, Akyildiz, Balandat, Spisak, Jain, Rabbat, and
  Hazelwood}{Wu et~al\mbox{.}}{2021}]%
        {wu:arxiv:2021}
\bibfield{author}{\bibinfo{person}{Carole-Jean Wu}, \bibinfo{person}{Ramya
  Raghavendra}, \bibinfo{person}{Udit Gupta}, \bibinfo{person}{Bilge Acun},
  \bibinfo{person}{Newsha Ardalani}, \bibinfo{person}{Kiwan Maeng},
  \bibinfo{person}{Gloria Chang}, \bibinfo{person}{Fiona~Aga Behram},
  \bibinfo{person}{James Huang}, \bibinfo{person}{Charles Bai},
  \bibinfo{person}{Michael Gschwind}, \bibinfo{person}{Anurag Gupta},
  \bibinfo{person}{Myle Ott}, \bibinfo{person}{Anastasia Melnikov},
  \bibinfo{person}{Salvatore Candido}, \bibinfo{person}{David Brooks},
  \bibinfo{person}{Geeta Chauhan}, \bibinfo{person}{Benjamin Lee},
  \bibinfo{person}{Hsien-Hsin~S. Lee}, \bibinfo{person}{Bugra Akyildiz},
  \bibinfo{person}{Maximilian Balandat}, \bibinfo{person}{Joe Spisak},
  \bibinfo{person}{Ravi Jain}, \bibinfo{person}{Mike Rabbat}, {and}
  \bibinfo{person}{Kim Hazelwood}.} \bibinfo{year}{2021}\natexlab{}.
\newblock \showarticletitle{Sustainable {AI: E}nvironmental Implications,
  Challenges and Opportunities}.
\newblock \bibinfo{journal}{\emph{CoRR}}  \bibinfo{volume}{abs/2111.00364}
  (\bibinfo{year}{2021}).
\newblock


\bibitem[\protect\citeauthoryear{Yin, Acun, Liu, and Wu}{Yin
  et~al\mbox{.}}{2021}]%
        {yin:2021:ttrec}
\bibfield{author}{\bibinfo{person}{Chunxing Yin}, \bibinfo{person}{Bilge Acun},
  \bibinfo{person}{Xing Liu}, {and} \bibinfo{person}{Carole{-}Jean Wu}.}
  \bibinfo{year}{2021}\natexlab{}.
\newblock \showarticletitle{TT-Rec: Tensor Train Compression for Deep Learning
  Recommendation Models}.
\newblock \bibinfo{journal}{\emph{CoRR}}  \bibinfo{volume}{abs/2101.11714}
  (\bibinfo{year}{2021}).
\newblock
\urldef\tempurl%
\url{https://arxiv.org/abs/2101.11714}
\showURL{%
\tempurl}


\bibitem[\protect\citeauthoryear{Zhang, Liu, Xie, Ktena, Tejani, Gupta, Myana,
  Dilipkumar, Paul, Ihara, Upadhyaya, Huszar, and Shi}{Zhang
  et~al\mbox{.}}{2020}]%
        {freq_hashing}
\bibfield{author}{\bibinfo{person}{Caojin Zhang}, \bibinfo{person}{Yicun Liu},
  \bibinfo{person}{Yuanpu Xie}, \bibinfo{person}{Sofia~Ira Ktena},
  \bibinfo{person}{Alykhan Tejani}, \bibinfo{person}{Akshay Gupta},
  \bibinfo{person}{Pranay~Kumar Myana}, \bibinfo{person}{Deepak Dilipkumar},
  \bibinfo{person}{Suvadip Paul}, \bibinfo{person}{Ikuhiro Ihara},
  \bibinfo{person}{Prasang Upadhyaya}, \bibinfo{person}{Ferenc Huszar}, {and}
  \bibinfo{person}{Wenzhe Shi}.} \bibinfo{year}{2020}\natexlab{}.
\newblock \showarticletitle{Model Size Reduction Using Frequency Based Double
  Hashing for Recommender Systems}. In \bibinfo{booktitle}{\emph{Fourteenth ACM
  Conference on Recommender Systems}} (Virtual Event, Brazil)
  \emph{(\bibinfo{series}{RecSys '20})}. \bibinfo{publisher}{Association for
  Computing Machinery}, \bibinfo{address}{New York, NY, USA},
  \bibinfo{pages}{521–526}.
\newblock
\showISBNx{9781450375832}
\urldef\tempurl%
\url{https://doi.org/10.1145/3383313.3412227}
\showDOI{\tempurl}


\bibitem[\protect\citeauthoryear{Zhao, Xie, Jia, Qian, Ding, Sun, and Li}{Zhao
  et~al\mbox{.}}{2020}]%
        {zhao:2020:distributedgpu}
\bibfield{author}{\bibinfo{person}{Weijie Zhao}, \bibinfo{person}{Deping Xie},
  \bibinfo{person}{Ronglai Jia}, \bibinfo{person}{Yulei Qian},
  \bibinfo{person}{Ruiquan Ding}, \bibinfo{person}{Mingming Sun}, {and}
  \bibinfo{person}{Ping Li}.} \bibinfo{year}{2020}\natexlab{}.
\newblock \showarticletitle{Distributed Hierarchical GPU Parameter Server for
  Massive Scale Deep Learning Ads Systems}. In
  \bibinfo{booktitle}{\emph{Proceedings of Machine Learning and Systems}}.
\newblock


\bibitem[\protect\citeauthoryear{Zhao, Hong, Wei, Chen, Nath, Andrews,
  Kumthekar, Sathiamoorthy, Yi, and Chi}{Zhao et~al\mbox{.}}{2019}]%
        {what_to_watch}
\bibfield{author}{\bibinfo{person}{Zhe Zhao}, \bibinfo{person}{Lichan Hong},
  \bibinfo{person}{Li Wei}, \bibinfo{person}{Jilin Chen},
  \bibinfo{person}{Aniruddh Nath}, \bibinfo{person}{Shawn Andrews},
  \bibinfo{person}{Aditee Kumthekar}, \bibinfo{person}{Maheswaran
  Sathiamoorthy}, \bibinfo{person}{Xinyang Yi}, {and} \bibinfo{person}{Ed
  Chi}.} \bibinfo{year}{2019}\natexlab{}.
\newblock \showarticletitle{Recommending What Video to Watch next: A Multitask
  Ranking System}. In \bibinfo{booktitle}{\emph{Proceedings of the 13th ACM
  Conference on Recommender Systems}} (Copenhagen, Denmark)
  \emph{(\bibinfo{series}{RecSys '19})}. \bibinfo{publisher}{Association for
  Computing Machinery}, \bibinfo{address}{New York, NY, USA},
  \bibinfo{pages}{43–51}.
\newblock
\showISBNx{9781450362436}
\urldef\tempurl%
\url{https://doi.org/10.1145/3298689.3346997}
\showDOI{\tempurl}


\bibitem[\protect\citeauthoryear{Zhou, Mou, Fan, Pi, Bian, Zhou, Zhu, and
  Gai}{Zhou et~al\mbox{.}}{2019}]%
        {zhou2019deep}
\bibfield{author}{\bibinfo{person}{Guorui Zhou}, \bibinfo{person}{Na Mou},
  \bibinfo{person}{Ying Fan}, \bibinfo{person}{Qi Pi}, \bibinfo{person}{Weijie
  Bian}, \bibinfo{person}{Chang Zhou}, \bibinfo{person}{Xiaoqiang Zhu}, {and}
  \bibinfo{person}{Kun Gai}.} \bibinfo{year}{2019}\natexlab{}.
\newblock \showarticletitle{Deep interest evolution network for click-through
  rate prediction}. In \bibinfo{booktitle}{\emph{AAAI conference on artificial
  intelligence}}, Vol.~\bibinfo{volume}{33}. \bibinfo{pages}{5941--5948}.
\newblock


\end{thebibliography}

\end{document}